\documentclass{article}

\usepackage{microtype}
\usepackage{graphicx}
\usepackage{booktabs} % for professional tables

\usepackage{hyperref}

\usepackage[preprint]{icml2025}

\usepackage{amsmath}
\usepackage{amssymb}
\usepackage{mathtools}
\usepackage{amsthm}
\usepackage{algorithmic}
\usepackage[capitalize,noabbrev]{cleveref}

\theoremstyle{plain}
\newtheorem{theorem}{Theorem}[section]

\theoremstyle{definition}

\theoremstyle{remark}
\newtheorem{remark}[theorem]{Remark}

\usepackage[textsize=tiny]{todonotes}

\usepackage{amsmath,amsfonts,bm}

\def\eqref#1{equation~\ref{#1}}
\def\Eqref#1{Equation~\ref{#1}}
\def\1{\bm{1}}

\DeclareMathAlphabet{\mathsfit}{\encodingdefault}{\sfdefault}{m}{sl}
\SetMathAlphabet{\mathsfit}{bold}{\encodingdefault}{\sfdefault}{bx}{n}

\def\sG{{\mathbb{G}}}

\def\sR{{\mathbb{R}}}

\usepackage{xcolor}
\usepackage{hyperref}
\usepackage{url}
\usepackage{graphicx}
\usepackage{wrapfig}
\usepackage{subcaption}
\newcommand{\eshdcpdag}{\text{ESHD}_{\text{CPDAG}}}
\newcommand{\fscorecpdag}{\text{F1-Score}_{\text{CPDAG}}}

\newcommand{\independent}{\perp\!\!\!\!\perp}
\newcommand{\ERfive}{\text{ER(5, 1)}}
\newcommand{\ERten}{\text{ER(10, 2)}}
\newcommand{\ERthirty}{\text{ER(30, 2)}}
\newcommand{\submethod}{NN-opt method}

\newcommand{\molko}[1]{\textcolor{orange}{\small [mo: #1]}}
\newcommand{\mg}[1]{\textcolor{magenta}{\small [mg: #1]}}
\newcommand{\asia}[1]{\textcolor{brown}{\small [jw: #1]}}
\newcommand{\piotrm}[1]{\textcolor{blue}{\small [pm: #1]}}
\newcommand{\piotrs}[1]{\textcolor{blue}{\small [ps: #1]}}
\newcommand{\kucil}[1]{\textcolor{violet}{\small [łk: #1]}}
\newcommand{\alicja}[1]{\textcolor{red}{\small [az: #1]}}
\newcommand{\mikolajm}[1]{\textcolor{teal}{\small [mm: #1]}}
\newcommand{\reb}[1]{\textcolor{red}{#1}}
\renewcommand{\molko}[1]{}
\renewcommand{\mg}[1]{}
\renewcommand{\asia}[1]{}
\renewcommand{\piotrm}[1]{}
\renewcommand{\piotrs}[1]{}
\renewcommand{\kucil}[1]{}
\renewcommand{\alicja}[1]{}
\renewcommand{\mikolajm}[1]{}
\renewcommand{\reb}[1]{#1}

\icmltitlerunning{The Performance
Limits of Neural Causal Discovery}

\begin{document}

\twocolumn[
\icmltitle{Since Faithfulness Fails: The Performance
Limits of Neural Causal Discovery}

% It is OKAY to include author information, even for blind
% submissions: the style file will automatically remove it for you
% unless you've provided the [accepted] option to the icml2025
% package.

% List of affiliations: The first argument should be a (short)
% identifier you will use later to specify author affiliations
% Academic affiliations should list Department, University, City, Region, Country
% Industry affiliations should list Company, City, Region, Country

% You can specify symbols, otherwise they are numbered in order.
% Ideally, you should not use this facility. Affiliations will be numbered
% in order of appearance and this is the preferred way.
\icmlsetsymbol{equal}{*}

\begin{icmlauthorlist}
\icmlauthor{Mateusz Olko}{equal,uw,ideas}
\icmlauthor{Mateusz Gajewski}{equal,ideas,pp}
\icmlauthor{Joanna Wojciechowska}{uw}\\
\icmlauthor{Mikołaj Morzy}{pp}
\icmlauthor{Piotr Sankowski}{uw,mimsolutions}
\icmlauthor{Piotr Miłoś}{uw,ideas,impan,deepsense}
% \icmlauthor{Firstname7 Lastname7}{comp}
%\icmlauthor{}{sch}
% \icmlauthor{Firstname8 Lastname8}{sch}
% \icmlauthor{Firstname8 Lastname8}{yyy,comp}
%\icmlauthor{}{sch}
%\icmlauthor{}{sch}
\end{icmlauthorlist}

\icmlaffiliation{uw}{Warsaw University, Warsaw, Poland}
\icmlaffiliation{pp}{Poznan University of Technology, Poznan, Poland}
\icmlaffiliation{ideas}{IDEAS NCBR, Warsaw, Poland}
\icmlaffiliation{impan}{Institute of Mathematics, Polish Academy of Sciences, Warsaw, Poland}
\icmlaffiliation{mimsolutions}{MIM Solutions, Warsaw, Poland}
\icmlaffiliation{deepsense}{deepsense.ai, Warsaw, Poland}

\icmlcorrespondingauthor{Mateusz Olko}{mateusz.olko@gmail.com}
\icmlcorrespondingauthor{Mateusz Gajewski}{mg96272@gmail.com}

% You may provide any keywords that you
% find helpful for describing your paper; these are used to populate
% the "keywords" metadata in the PDF but will not be shown in the document
\icmlkeywords{Machine Learning, ICML}

\vskip 0.3in
]

% this must go after the closing bracket ] following \twocolumn[ ...

% This command actually creates the footnote in the first column
% listing the affiliations and the copyright notice.
% The command takes one argument, which is text to display at the start of the footnote.
% The \icmlEqualContribution command is standard text for equal contribution.
% Remove it (just {}) if you do not need this facility.

%\printAffiliationsAndNotice{}  % leave blank if no need to mention equal contribution
\printAffiliationsAndNotice{\icmlEqualContribution} % otherwise use the standard text.
\vspace{-1.5em}
\begin{abstract}

\looseness=-1 Neural causal discovery methods have recently improved in terms of scalability and computational efficiency. However, our systematic evaluation highlights significant room for improvement in their accuracy when uncovering causal structures. We identify a fundamental limitation: \textit{neural networks cannot reliably distinguish between existing and non-existing causal relationships in the finite sample regime}. Our experiments reveal that neural networks, as used in contemporary causal discovery approaches, lack the precision needed to recover ground-truth graphs, even for small graphs and relatively large sample sizes. Furthermore, we identify the faithfulness property as a critical bottleneck: (i) it is likely to be violated across any reasonable dataset size range, and (ii) its violation directly undermines the performance of neural discovery methods. These findings lead us to conclude that progress within the current paradigm is fundamentally constrained, necessitating a paradigm shift in this domain.

% To investigate further, we use strong-faithfulness property to characterize benchmark datasets, and show that datasets violating this property present substantially greater challenges to causal discovery methods. Critically, distributions adhering to strong faithfulness become exponentially rare as graph size increases, posing a severe obstacle to scaling neural causal discovery methods.

\end{abstract}
\vspace{-1.7em}
\section{Introduction}

% \begin{wrapfigure}{r}{0.4\textwidth}
% \vspace{-1cm}
%   \begin{center}
%     \includegraphics[width=1\linewidth]{imgs/faith_main_with_numbers.pdf}
%   \end{center}
%   \vspace{-0.5cm}
%   \caption{Neural causal discovery methods suffer from inherent performance limit due to violation of faithfulness assumption, but there is still room for improvement. \reb{Values computed for ER(5,1) class of graphs. See Sections~\ref{sec:benchmark},~\ref{sec:nn_opt}.}}
%     \vspace{-0.4cm}
% \end{wrapfigure}

% \mg{Estimated time EOD 27.01}

\looseness=-1 Causal discovery is essential to scientific research, driving a growing demand for machine learning methods to support this process. Despite the development of several neural causal discovery methods in recent years~\citep{dcdi,dibs,bayesdag,sdcd}, their performance remains insufficient for real-world applications, particularly in fields like medicine and biology~\citep{causality_medical_imaging, faithful_biology}. Furthermore, these methods are usually evaluated using datasets, which vary between studies, obscuring the overall picture and making assessment of advancements difficult.

In response to this challenge, we introduce a unified benchmarking protocol for neural causal discovery methods. Specifically, we use identical datasets, tune hyperparameters consistently, and use a standardized functional approximation across all methods. Through this systematic evaluation, we uncover a concerningly low level of performance. This raises a fundamental question: Why do these methods struggle to recover the ground-though causal structure even when the data is generated synthetically with all theoretical assumptions fulfilled and the dataset is large? %Understanding this limitation is crucial for the advancement of the field.

We identify a fundamental limitation: \textit{neural networks cannot reliably distinguish between existing and non-existing causal relationships in the finite sample regime}. This happens due to unavoidable approximation and estimation errors. We observe minimal improvements when scaling either the data or the network size. To make matters worse, the difficulty of the problem escalates rapidly with increasing graph size. Based on these observations, we pose a hypothesis that \emph{tangible progress cannot be achieved within the current neural discovery paradigm}. At a technical level, we pinpoint the faithfulness property as a critical bottleneck.\piotrs{ja chyba napisałbym z jedno zdanie co to jest faithfulness} This assumption proves to be highly brittle and is likely violated across any reasonable dataset size range.

We believe these limitations highlight the need for a paradigm shift rather than incremental improvements in architecture or optimization techniques. 
% In Section \piotrm{…}, we discuss viable directions for future research, such as developing better approximators or exploring alternative structure discovery objectives.

In more details, our contributions are as follows:
% \molko{We need to reiterate over contributions}
\vspace{-1em}
\begin{enumerate}
    \itemsep-0.1em 
    \item In a carefully controlled experiment, we demonstrate that neural networks, as used by causal discovery approaches, lack the precision required to recover ground-truth causal graphs, even for small graphs and relatively large sample sizes. We attribute this problem to estimation error stemming for finite dataset size.
    \item We propose practical method of approximation of $\lambda$-strong faithfulness property for nonlinear datasets. We experimentally demonstrate how proportion of $\lambda$-unfaithfull distributions scales with the size and density of graphs from Erdos-Renyi class. 
    \item We provide results of unified benchmark for neural causal discovery methods and show that their performance correlates with $\lambda$-strong faithfulness property of the datasets. 
\end{enumerate}

\section{Background and Related Work} \label{sec:background}

\paragraph{Structural Causal Models (SCMs)} \label{par:scm} Causal relationships are commonly formalized \citep{pearl} using SCMs, which represent causal dependencies through a set of structural equations. For a directed acyclic graph (DAG) $G = (V, E)$, an SCM is defined by %a set of equations
\begin{equation}
 X_i = f_i(Pa_i, U_i),
\end{equation}
where $X_i$ is a random variable at vertex $i \in V$, $f_i \colon \sR^{|Pa_i| + 1} \rightarrow \sR$ is a function, $Pa_i$ denotes the set of parents of $i$ in $G$, and $U_i$ is an independent noise. % term associated with $X_i$.
In this work, we assume
\textit{additive noise} SCMs, also referred to as \textit{additive noise models} (ANM), viz., 
\begin{equation}\label{eq:additive_scm}
f_i(Pa_i, U_i) = g_i(\mathit{Pa}_i)+ U_i
\end{equation}

for some $g_i \colon \sR^{|Pa_i|} \rightarrow \sR$. An SCM defines a joint distibution $P$ over the set of random variables $\{X_i\}_{i \in V}$.
%An SCM defines a joint distibution $P$ over the set of random vairables $\{X_i\}$.

\paragraph{Causal Discovery} 
Causal discovery aims to uncover causal structure $G$ of a SCM based on data sampled from the joint distribution $P$.
% \piotrm{17.02 <-- make the first sentence more precise}. 
However, in general the solution can only be identified up to a Markov Equivalence Class (MEC), the set of DAGs encoding the same conditional independencies~\citep{Verma1990EquivalenceAS}. 
% However, in general the unique solution cannot be identified from the observational data only; instead, one can only identify the structure up to a Markov Equivalence Class (MEC), the set of DAGs that encode the same conditional independencies~\citep{Verma1990EquivalenceAS}. 

While there is a line of classical methods like PC~\citep{pc_sprites} or GES~\citep{chickering2020statistically}, those algorithms work well for linear relationships and moderate-sized problems, they struggle in more challenging scenarios. 
% \piotrm{17.02 <-- not sure what this paragraph is for?}
% \paragraph{Neural causal discovery}
Neural causal discovery methods have emerged as a promising direction, offering class of methods that could potentially handle nonlinear relationships and provide computational efficiency, and scalability in both data size and number of variables. These methods, use neural networks as functional approximators and continuous optimization techniques to learn both the graph structure and causal mechanisms simultaneously. They allow handling challenging scenarios and therefore should be applicable in many real-wold problems. 

Evaluating causal discovery methods presents unique challenges. In real-world applications, ground truth causal structures are typically unknown or can only be partially elicited from domain experts with inherent noise and bias. This fundamental limitation necessitates the use of synthetic data for evaluations, as they provide controlled environments essential for understanding the methods' capabilities and limitations.
\label{sec:related_work}

\paragraph{Recent Developments in Neural Causal Discovery}
\label{par:score_based}

%\mg{28.01 new version}

% We evaluated recent neural causal discovery methods such as: NO-TEARS, NO-BEARS, NO-CURL, GRAN-DAG, SCORE, DAGMA, DCDFG, DCDI, DiBS, BayesDAG, SDCD. 
We would like to highlight four recent approaches to neural causal discovery: DCDI~\citep{dcdi}, SDCD~\citep{sdcd}, DiBS~\citep{dibs}, and BayesDAG~\citep{bayesdag}, which in our view effectively represent the major developments in neural causal discovery from  past years. DCDI represents the evolution of NO-TEARS-based methods like GRAN-DAG~\citep{notears, grandag}, improving upon the original by separating structural and functional parameters and incorporating interventional data. SDCD unifies and advances various acyclicity constraints proposed in methods like NO-BEARS~\citep{nobears} and DAGMA~\citep{dagma}, demonstrating superior performance compared to SCORE~\citep{score} and DCDFG~\citep{dcdfg}. For Bayesian approaches, we chose DiBS, which incorporates NO-TEARS regularization in its prior, and BayesDAG, which builds on NO-CURL's DAG parametrization~\citep{nocurl} using MCMC optimization. 

All these approaches use a continuous representation of the graph structure, enforcing a differentiable acyclicity constraint to ensure the result is a DAG. The primary objective is to maximize $\log p_\theta(X|\mathcal{G})$, that is the log-likelihood of the data given the graph while incorporating regularization terms to control graph complexity.
The discovery procedure comprises two parts: fitting functional approximators and structure search, which are usually done in parallel.

\paragraph{The Faithfulness Property} 
% \piotrm{17.02 condition vs assumption}
% \molko{Lambda strong property, assumption or condition? Can we skip faithfulness?}
%The properties of distributions generated by Bayesian Networks with relation to the underlying graph have been extensively studied. The most common assumption relating the two is the faithfulness assumption. 
The so-called faithfulness property is a fundamental assumption commonly used by causal discovery methods \citep{pearl, dcdi}. 
% It states that $d$-separation in a DAG holds if and only if conditional independencies in the distribution:
\begin{equation}
    \forall_{a,b \in V} \forall_{S \subseteq V \setminus \{a ,b\}}X_a \perp\!\!\!\perp X_b | X_S \iff a \perp_\mathcal{G} b | S.
\end{equation}
where  $\perp\!\!\!\perp$ denotes conditional independence and $\perp_\mathcal{G}$ denotes $d$-separation. For more information on d-separation please refer to Appendix~\ref{appendix:d_separation}.
% \piotrm{17.02 <-- this definition needs to be written properly. What is $\perp_\mathcal{G}$}
The faithfulness property can be violated when multiple causal paths cancel each other (see example in Appendix~\ref{appendix:faith_example}). While unfaithful distributions associated with a given DAG have measure zero in the space of possible distributions~\citep{boeken2025bayesiannetworkstypicallyfaithful}, we can encounter distributions where causal relationships are arbitrarily weak. Without infinite samples, such relationships remain undetectable. 

This observation motivated~\citealp{zhang2003faithfulness} to introduce the $\lambda$-strong faithfulness assumption. % for linear systems. 
A distribution $P$ is said to be $\lambda$-strong faithful to a DAG $\mathcal{G}$ if:

\begin{equation}\label{eq:lambda_faith}
    \forall_{a,b \in V} \forall_{S \subseteq V \setminus \{a ,b\}} |\rho_P(X_a, X_b | X_S)| > \lambda \iff a \not\perp_\mathcal{G} b | S,
\end{equation}

where $\rho_P(X_a, X_b | X_S)$ denotes a partial \mg{20.02 should we write 'linear'?}\molko{21.02 I think they do not really specify that it is linear. Though maybe from the context it can be deduced? I would vote for not specifying unless we have to} correlation coefficient and $\lambda \in (0,1)$\footnote{When $\lambda=0$, $\lambda$ this reduces to the standard faithfulness assumption.}.  For linear systems, this assumption ensures uniform consistency of the PC algorithm with $\lambda \propto 1 / \sqrt{n}$ \mg{can we explain it in one sentence? What doeas it mean? Because I am not sure }where the number of nodes $p = |V|$ is fixed and sample size $n \rightarrow \infty.$ Therefore, $\lambda$ can serve as a notion of the difficulty of the causal discovery task. 

% Additionally, \citet[Section 3.2]{zhang2003faithfulness} note that when a distribution is nearly unfaithful, small changes in parameters can disrupt its dependence structure. In this sense, $\lambda$ in the $\lambda$-Strong-Faithfulness condition provides a rough measure of how sensitive a distribution is to such changes. \piotrm{17.02 <-- do we need this? It hard to parse, and not informaiive?}

Notably ~\citep{geometry_of_faithfulness} proved that, in case of linear SCMs, for any fixed $\lambda > 0$, the fraction of $\lambda$-strong faithful distributions decreases exponentially with graph size and density, suggesting fundamental limitations in causal discovery on large graphs. \piotrm{17.02 rephrase. } \piotrm{17.02 important: how does this depend on $\lambda$.}

While above mentioned work provide theoretical results for linear SCMs and the PC algorithm. So far little has been shown regarding nonlinear functions and contemporary neural network approaches. We aim to breach the gap with our experimental contributions.

For a fixed distribution $P$ and associated with it graph $\mathcal{G}$ there can be multiple $\lambda$ that satisfy the Equation~\ref{eq:lambda_faith}. Therefore, in the remainder of the paper we will denote $\lambda$ as the maximal threshold satisfying the equation. Additionally, as we are working with nonlinear data, we use Spearman correlation coefficient, which we will denote $\rho_P$ with a slight abuse of notation. Specifically, for a given distribution $P$ associated with graphs $\mathcal{G}$ we define:
\begin{multline}\label{eq:lambda_def}
    \lambda = \max\{t: \forall_{a,b \in V} \forall_{S \subseteq V \setminus \{a ,b\}} \\ |\rho_P(X_a, X_b | X_S)| > t \iff a \not\perp_\mathcal{G} b | S\}.
\end{multline}

% \piotrm{17.02 I'm mathematically biased. I'd prefer to see $\forall$ etc}

% \molko{New narration:
% \begin{enumerate}
%     \item 1) We begin nonlinear case analysis by providing method to apprximate lambda. 2) The $\hat{\lambda}$ behaves as expected and described by Uhler on non linear (connected) ER graphs.
%     \item 1) in order to analyse convergeence rate of ncd methods we design a robust approach that minimizes aproximation errors 2) We justify robustness (text, no exps)
%     3) we show that number of required samples depends on lambda
%     4) we provide case study of estimation error, we show tnere is statistically signifficant error in score modelling (we atrribute it to limited data)
%     5) we show that the method improves signifficantly when provided larger dataset
%     \item We conclude that because number of lambda faithfull distibutions vanishes quickly and lambda is connected to the number of samples required to recover the true graph the hight-dimensional effectivenes of the current ncd paradigm seems rather unreaslitic (unless real world data has substantially different characteristic that random NNs)
%     \item (no changes reqruied) We show that the above metioned observations apply to contemporary NCD methods on medium sized graphs. 2) Even medium sized graphs cause problems which highlights the limitaions for scaling this paradighm to larger graphs.
% \end{enumerate}
% }

\section{Main Result}
\label{sec:main}
\piotrm{17.02 more sexy title?}

% \piotrm{17.02 The main goal of this section is to show that the causal discovery problem, as stated above, if fundamentally hard. To this end we build a framework that allows us to measure the difficulty of the problem (Section 3.1) and show that the difficulty is connected to the number of samples required to recover the true graph (Section 3.2). Our farmework mimicks the linear case described in the previou secion.}

% This paper investigates the relationship between $\lambda$ and neural causal discovery in nonlinear datasets.
% In Section~\ref{sec:difficulty}, we introduce a method for approximating $\lambda$ in finite nonlinear datasets and demonstrate that $\lambda$-strong faithful nonlinear datasets based on ER graphs exhibit similar behavior to the linear datasets described by \citet{geometry_of_faithfulness}, vanishing rapidly as graph density and size increase. In Section~\ref{sec:nn_opt}, we propose a conceptually simple approach to minimizing approximation errors in neural causal discovery and establish a relationship between $\lambda$ and the number of samples required for successful structure recovery. 
The main goal of this section is to show that the causal discovery problem, as stated above, is fundamentally hard. To this end we build a framework that allows us to measure the difficulty of the problem and show it grows quickly with the size and density of the underlying graph (Section 3.1). Than, we demonstrate that the difficulty is connected to the number of samples required to recover the true graph (Section 3.2). Our framework mimics the linear case described in the previous section.
Section~\ref{sec:case_study} presents a case study illustrating how small sample sizes affect structure scores, highlighting the severity of the problem. Finally, in Section~\ref{sec:benchmark}, we evaluate contemporary neural causal discovery methods and analyze their performance in relation to the problem difficulty measure.

% We present our findings in the following way: In Section~\ref{sec:difficulty}, we describe a method of approximating $\lambda$ for finite nonlinear datasets, and show that lambda-strong faithful nonlinear datasets based on ER graphs behave similarly to the linear datasets described by \citet{geometry_of_faithfulness} i.e. they vanish quickly with increasing density and size of graphs; In Section~\ref{sec:nn_opt}, we propose a conceptually simple method that minimizes approximation errors in neural causal discovery and relate $\lambda$ of the dataset with the amount od samples required for successful structure discovery; In section~\ref{sec:case_study}, we provide a case study which visualizes how small sample dataset influences the scores  assigned to the structures and signifies the severity of the problem. Later, in Section~\ref{sec:benchmark}, we present experimental evaluation of contemporary neural causal discovery methods and relate their performance to lambda-strong faithfulness property of the datasets.

\begin{figure*}[htbp]
\centering
\begin{subfigure}[t]{0.48\textwidth}
        \centering
        \includegraphics[width=\textwidth]{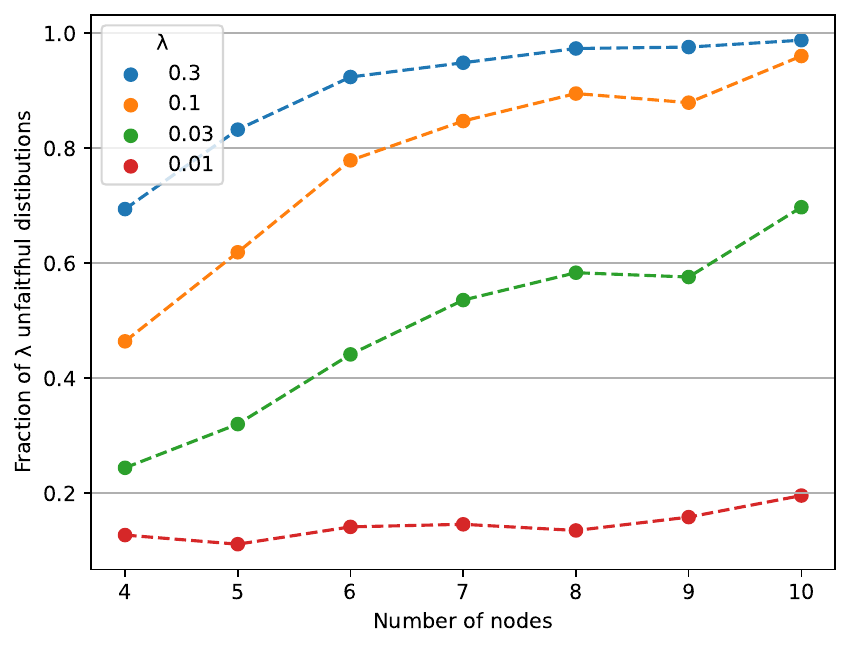}
    \caption{}
\label{fig:geom_fixed_nodes}
\end{subfigure}
~
\begin{subfigure}[t]{0.48\textwidth}
        \centering
        \includegraphics[width=\textwidth]{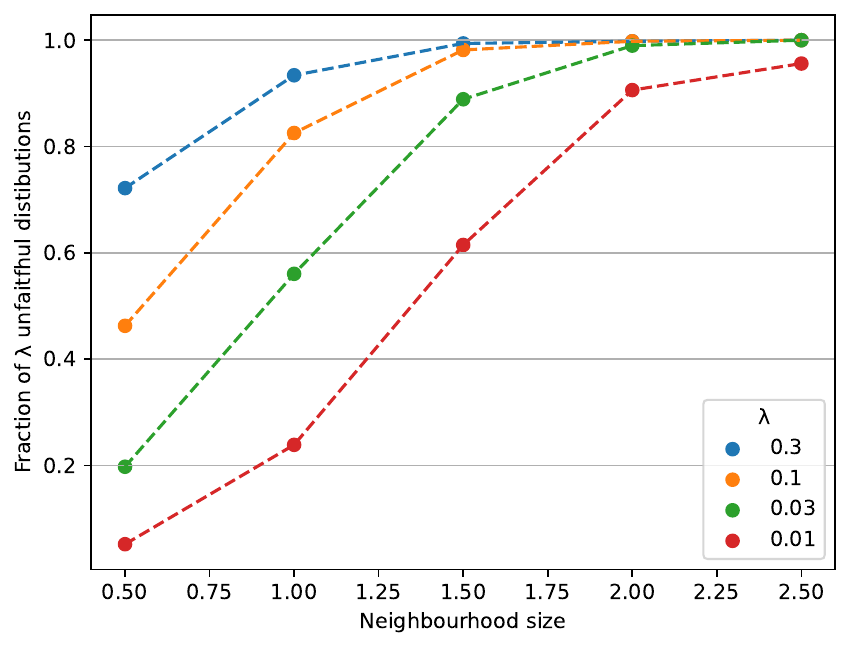}
    \caption{}
\label{fig:geom_fixed_density}
\end{subfigure}
\caption{(a) Estimated fraction of $\lambda$-unfaithful distributions for Erdos-Renyi graphs with various number of nodes. %Colored lines correspond to specific values of $\lambda$ 
(b) Estimated fraction of $\lambda$-unfaithful distributions for Erdos-Renyi graphs with 6 nodes and varying density. Colored lines correspond to specific values of $\lambda$.}

\end{figure*}

\subsection{The difficulty of a problem grows quickly with graph size and density}
\label{sec:difficulty}
% \molko{Dictionary:
% \begin{itemize}
%     \item We approximate lambda \textbf{not estimate}.
%     \item We propose a \textbf{measure} of difficulty of disto, not metric.
% \end{itemize}}

% In Section 3, we observe that convergence depends on the properties of the distribution. To better understand the severity of the problem we seek to quantify those properties and estimate how often we can expect to encounter adversarial conditions. In this section we propose a practical method ot approximate a measure to difficulty of the distribution.

% Recall from Section~\ref{sec:background} that $\lambda$ in the $\lambda$-strong faithfulness assumption can serve as a notion of difficulty of a dataset in finite data. We aim at approximating $\lambda$ as it can provide insights into the inherent difficulty of learning causal structure from a given distribution. 

% \mg{intro: Since $\lambda$ is connected to the difficulty of causal discovery, we would like to have a method to approximate it for finite nonlinear datasets.}
We aim to measure the difficulty of causal discovery given non-linear dataset $\mathcal{D}$ and associated with it structure $\mathcal{G}$ expressed by the $\lambda$ parameter.
Theoretically, $\lambda$ could be determined by computing partial correlations for all variable pairs across all conditioning sets and identifying a threshold that separates conditionally d-separated from d-connected nodes (see \eqref{eq:lambda_def}). However, due to irreducible errors in correlation estimation from finite data and the presence of small true correlation values,  , it is infeasible to establish a precise threshold in practice. To address this, we approximate $\lambda$ by selecting the threshold that maximizes the F1-score of a prediction. 
\piotrm{17.02 rewrite. Do we have any strong evidence that using \eqref{eq:lambda_def} would not work?} \molko{We have: the noise from spearman estimation is large compared to the values of spaerman correaltion that we are interested with}

\paragraph{Effective approximation of $\lambda$.}
Specifically, we use partial Spearman correlation, classifying node pairs as d-separated if their conditional correlation coefficient, computed from the finite dataset, falls below a given threshold. 

We then define $\hat{\lambda}$ as the threshold that optimizes the F1-score of this classification. Formally, for dataset $\mathcal{D}$ and associated with it DAG $\mathcal{G}$:   

\begin{equation}
% \textit{DeFaith} = \text{argmin}_{t \in \mathbb{R}} \left\{ 
%     \sum_{\substack{(a,b,S): \\ \text{d-sep}(a,b|S)}} \mathbb{1}\{|\rho_{ab|S}| > t\} +
%     \sum_{\substack{(a,b,S): \\ \neg\text{d-sep}(a,b|S)}} \mathbb{1}\{|\rho_{ab|S}| \leq t\} \right\}
\hat{\lambda} = \text{argmax}_t \text{F}_1(t, \mathcal{D}, \mathcal{G})
\end{equation}
\begin{equation*}
    \text{F}_1(t, \mathcal{D}, \mathcal{G}) = 2 \cdot \frac{\text{Precision}(t, \mathcal{D}, \mathcal{G}) \cdot \text{Recall}(t, \mathcal{D}, \mathcal{G})}{\text{Precision}(t, \mathcal{D}, \mathcal{G})+ \text{Recall}(t, \mathcal{D}, \mathcal{G})} \\
\end{equation*}
\begin{align*}
    \text{Precision}(t, \mathcal{D}, \mathcal{G}) &= \frac{\text{TP}(t, \mathcal{D}, \mathcal{G})}{\hat{\text{P}}(t, \mathcal{D}, \mathcal{G})} \\
    \text{Recall}(t, \mathcal{D}, \mathcal{G}) &= \frac{\text{TP}(t, \mathcal{D}, \mathcal{G})}{\text{P}(t, \mathcal{D}, \mathcal{G})}
\end{align*}
\begin{align*}
    \text{P}(t, \mathcal{D}, \mathcal{G}) &= \#\{a, b, S: a \independent_G b | S\} \\
    \hat{\text{P}}(t, \mathcal{D}, \mathcal{G}) &= \#\{a, b, S: |\rho_D(a, b | S)| < t\} \\
    \text{TP}(t, \mathcal{D}, \mathcal{G}) &= \#\{a, b, S: |\rho_D(a, b | S)| < t \text{ and } a \independent_G b | S\}
\end{align*}
where $\rho_D(a, b | S)$ denotes the partial Spearman correlation coefficient computed from dataset $\mathcal{D}$, and $a \independent_G b | S$ denotes $d$-separation in graph $\mathcal{G}$. We analyze the consistency of $\hat\lambda$ computation in the Appendix~\ref{sec:hat_lambda_errors}.
\begin{remark}
Note, that when data size increases estimation errors in the partial Spearman correlation $\rho_D(a, b | S)$ decrease. Thus, in the limit of the data $\hat{\lambda}$  converges to $\lambda$.
\end{remark}

\paragraph{Fraction of $\lambda$-strong distributions}

% Since we showed that the $\lambda$-strong property of a distribution have an impact on the difficulty of the dataset, we are interested in knowing how much does $\lambda$-strong faithfulness change with the graph density and number of nodes. 

In case of linear datasets, the relation between the fraction of sampled $\lambda$-strong distribution and the graph density and number of nodes has been extensively studied by~\citet{geometry_of_faithfulness}. It was proved, for several types of connected graphs, that the fraction of $\lambda$-strong faithful distributions decreases exponentially. We provide analogous experimental results for connected nonlinear graphs from the Erdos-Renyi class. 

In Figures~\ref{fig:geom_fixed_nodes} and~\ref{fig:geom_fixed_density} we plot how fraction of $\lambda$-faithful distributions changes with the number of nodes and the density of the graph respectively. Our experimental results align with theoretical study on linear datasets. We observe that for a fixed $\lambda$ and number of nodes, the proportion of distributions that are not $\lambda$-strong decreases with the expected neighborhood size, similar observation can be made when we increase the number of nodes in a graph (while keeping the graph connected). Even for relatively small and sparse graphs ($6$ nodes and expecte neighbourhood size $1$) only about half of distribution is faithful for $\lambda = 0.03$ and $20\%$ for $\lambda = 0.1$. 
% \piotrm{17.02 IMPORTANT: this is crucial to go beyond vague statements like: 'very fast'}
% \mg{19.02 rewritten}

\begin{figure*}[htbp]
\centering
\begin{subfigure}[t]{0.48\textwidth}
    \centering
    \includegraphics[width=\textwidth]{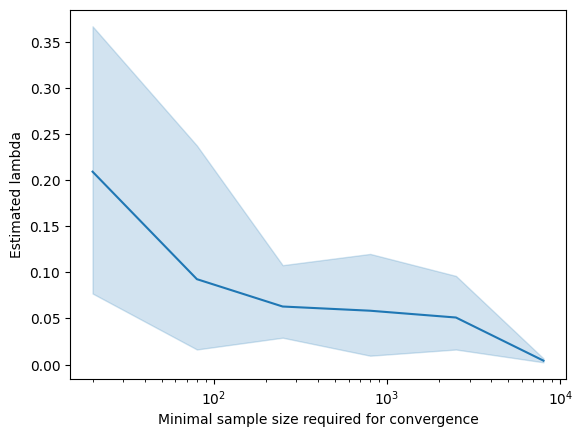}
    \caption{}
    \label{fig:converge}
\end{subfigure}
~
\begin{subfigure}[t]{0.48\textwidth}
    \centering
    \includegraphics[width=\textwidth]{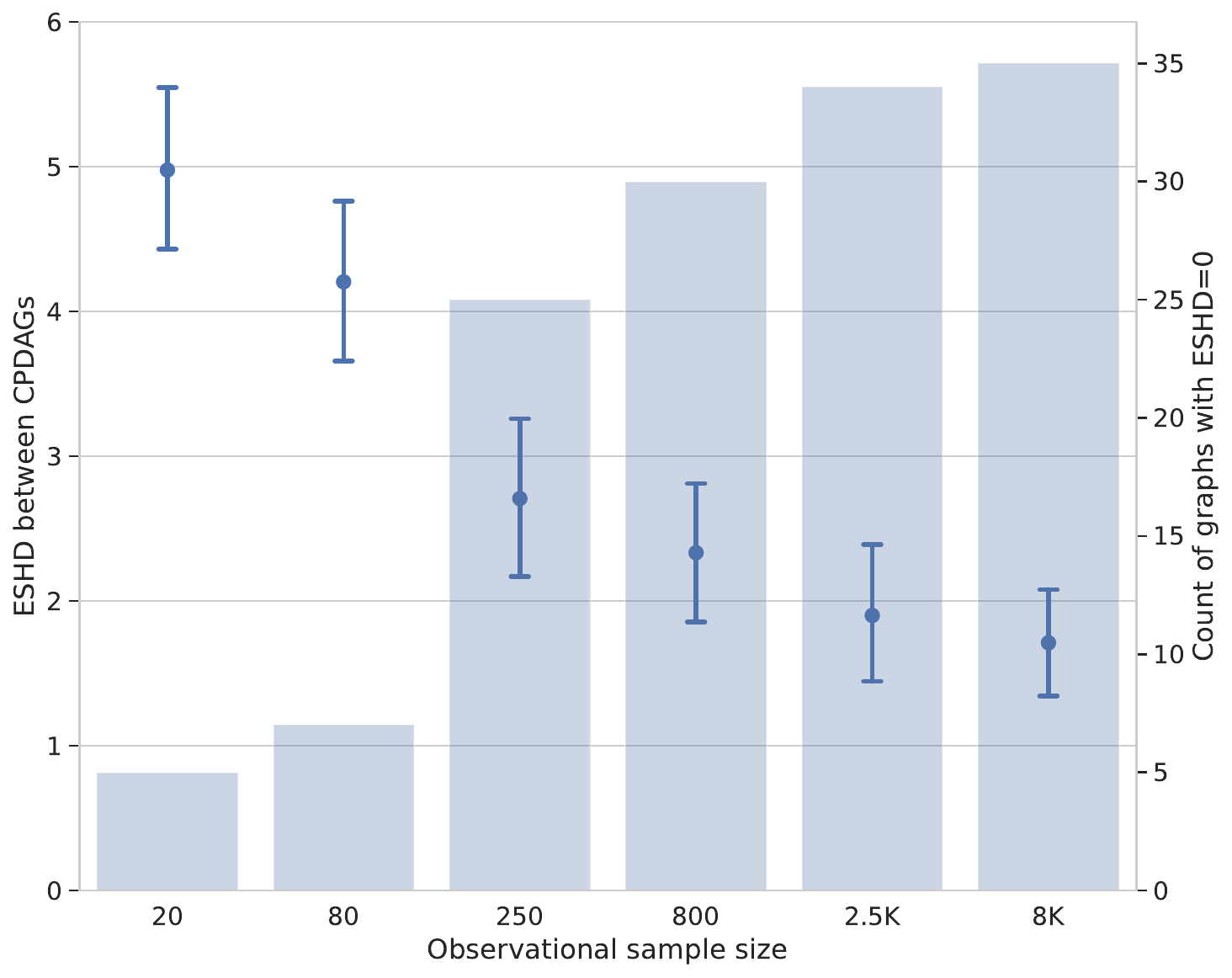}
    \caption{}
    \label{fig:naive_quantile}
\end{subfigure}
\caption{(a) Relation of sample needed to for a distribution to converge and the $\hat{\lambda}$.
(b) Comparison of the performance of \submethod{}  depending on data size. Averaged over 90 samples. For definition of $\eshdcpdag$ please refer to Section~\ref{sec:benchmark}.}
\vspace{-1em}
\end{figure*}
\subsection{Lower $\hat{\lambda}$ requires larger sample size}\label{sec:nn_opt}

% \molko{Naming conventions. Lets try to stick to that:
% \begin{itemize}
%     \item Datasets are \textbf{finite} but \textbf{large}, they are not limited.
%     \item NCD methods use neural-network functional approximators. They do not use neural network models.
%     \item Data is generated using a functional-model e.g. neural network.
%     \item how do we call $\lambda$ property?
% \end{itemize}}

% In this section, we investigate neural networks' ability to differentiate between structures based on the negative log-likelihood of the data. \molko{Change that}

% Specifically, we show that with r
% % Specifically, we show that even using 8K samples for a rather small Bayesian Networks it is impossible to train neural-network-based approximator to select ground truth structure using the score described in Section~\ref{sec:background}. 
% We claim that the problem is caused by estimation errors due to finite sample dataset.
% \piotrm{30.01 can we remove anything from this section? It is quite loaded}. 
% % \piotrm{30.01 be careful with stating that dataset is big in one place and limited in another.}
We investigate how $\lambda$ affects convergence rate of causal discovery algorithms with neural-network approximators. To be able to draw conclusions about impact of data sample on the accuracy of causal discovery
we design a conceptually simple neural network based causal discovery methods that minimizes approximation errors. We evaluate it in a controlled setting that enables a comprehensive analysis to identify bottlenecks in the neural causal discovery process.

\paragraph{Robust approximation approach}
% Our goal is to approximate the regularized log-likelihood score, a common objective function in neural causal discovery. For a graph $G$ with nodes set $V$ the log-likelihood of the dataset $D = \{x^{(1)}, ...,x^{(n)}\}$, consisting of vectors $x^{(j)} = \{x^{(j)}_i\}_{i \in V}$ can be expressed as the sum of  log-likelihoods of individual nodes $i \in V$:

% \begin{equation}\label{eq:score}
%     L(G) = \sum_{i \in V} \sum_{x \in D} \log p(x_i|\text{Pa}_i(G)) 
% \end{equation}

Our goal is to approximate the regularized log-likelihood score, a common objective function in neural causal discovery. For a graph $\mathcal{G}$ with nodes set $V$ the log-likelihood of the dataset $D = \{X_i\}_{i \in V}$ given a structure $G$ over set of nodes $V$ can be expressed as the sum of  log-likelihoods of individual vectors $X_i$:

\begin{equation}\label{eq:score}
    L(\mathcal{G}) = \sum_{i \in V} - \log p(X_i|\text{Pa}_i(\mathcal{G})) 
\end{equation}

Overall, the score can be written as:

% \begin{equation}\label{eq:score}
%     S(G) = \sum_{i \in V} - \log f_{\theta_{i,Pa_i(G)}, \sigma_i}(X_i, X_{\text{Pa}_i(G)}) + \gamma|G|
% \end{equation}
\begin{equation}\label{eq:score}
    S(\mathcal{G}) = \sum_{i \in V} - \log f(X_i; X_{\text{Pa}_i(G)};\theta_{i,Pa_i(\mathcal{G})}) + \gamma|\mathcal{G}|
\end{equation}
where $\theta_{i,Pa_i(\mathcal{G})}$ denotes the set of a parameters of neural-network-based density function $f$ of $X_i$ given set of parents of node $i$ $Pa_i(\mathcal{G})$.
% and $\sigma_i$ denotes the trained variance for the node $i$.
We would like to stress the fact that separate set of parameters $\theta$ is trained for each node $i \in V$ and for each possible parent set $Pa_i(\mathcal{G}) \subseteq V \setminus \{i\}$ . 

% Our goal is to approximate the regularized log-likelihood score, a common objective function in neural causal discovery. The log-likelihood of the data $\{X_1, ...,X_n\}$ given a structure $G$ can be expressed as the sum of  log-likelihoods of individual  vectors $X_i$:

% \begin{equation}\label{eq:score}
%     s(G) = \sum_{i \in V} - \log p(X_i|f_i(\text{pa}_i(G)), \sigma_i) + \gamma |G|.
% \end{equation}

% \mg{I think there is something not quite right with this equation $f_i(pa_i(G)$}

% Thus, to calculate $s(G)$ it suffices to compute the maximum likelihood estimate for each node \( i \in V\) and each possible set of parents \(S \subset V \setminus \{i\}\). To achieve this, for every $i$ and $S$, we train a neural network to approximate the function \(f_i(S)\) and optimize a single parameter representing \(\sigma_i\).
% For any DAG \(G \in \mathcal{G}_5\), the score \(s(G)\) is then computed by summing the corresponding likelihood contributions. 

To enhance robustness and quantify uncertainty, we employ random seed bootstrapping: the training procedure is repeated \(N\) times, and the final likelihood estimate is obtained as the mean over all \(N\) neural network approximations:

\begin{equation}\label{eq:score}
    \hat{S}(\mathcal{G}) = \sum_{i \in V} - \frac{1}{N} \sum_{n=1}^{N} \log f(X_i; X_{\text{Pa}_i(\mathcal{G})};\theta_{i,Pa_i(\mathcal{G})}^{(n)} + \gamma|\mathcal{G}|
\end{equation}

% \begin{equation}
% \hat{s}(G) = \sum_{i \in V} - \frac{1}{N} \sum_{n=1}^{N} \log p(X_i | f_i^{(n)}(\text{pa}_i(G)), \sigma_i^{(n)}) + \gamma |G|.
% \end{equation}

This approach helps mitigate estimation errors and provides a notion of uncertainty in the score computation.

\begin{remark}
Assuming: that the functional approximators are expressive enough, $\gamma$ is small enough and the distribution is faithful to the graph, the DAG that minimizes the score (\Eqref{eq:score}) should belong to the same Markov Equivalence Class as the graph used for data generation~\citep{dcdi}.
\end{remark}

The described procedure is meant to reduce the approximation error as much as possible at the expense of computational cost. This let us to disentangle the estimation and approximation effects, in particular we claim that the approximation error is negligible. To recap, we have:
\begin{enumerate}
    \item The neural network approximator and the data-generating model belong to the same class. % of models, ensuring consistency in function representation.
    \item We evaluate multiple network architectures and select the optimal one (see Table~\ref{tab:opt_models} in Appendix~\ref{app:nn_opt}). % balancing flexibility and approximation error .
    % \item The approximator is trained following best practices, including early stopping based on validation performance to prevent overfitting. \piotrm{31.01 this seems like saying: we do not have bugs ;)}
    \item We use an ensemble approach over multiple random seeds, which not only improves robustness but also provides bootstrap confidence interval for the approximation error.
\end{enumerate}

\begin{remark}
Notice that training and evaluation procedure described above can be considered as causal discovery method. The discovery is conducted in two steps: neural network training and exhaustive evaluation of space of DAGs. The method is not practically useful, but can illustrate the influence of estimation and approximation errors on causal discovery process. Since, we removed graph search component the method can also be viewed as upper-bound on existing differentiable approaches to causal discovery relying on similar neural-network approximators.
\end{remark}

\paragraph{Synthetics Causal Data} We generate synthetic data with a known ground-truth causal structure. We consider causal DAGs with only five nodes $V=\{1, \ldots, 5\}$. We generate these DAGs using the Erdos-Renyi model with the expected number of $5$ edges. The functional relations between nodes are modeled by randomly initialized MLPs with two hidden layers of size $8$. Additionally, we assume additive Gaussian noise: $U_i \sim \mathcal{N}(0, \sigma_i)$, where $\sigma_i$ may depend on $i \in V$. For more details refer to Appendix~\ref{appendix:dataset_generation}.

We generate 90 synthetic datasets and evaluate causal discovery approach described in previous section. We experiment with various neural-network sizes and select the best performing one which happens to be identical as used to generate the data (see Appendix~\ref{app:nn_opt}). The ensemble size used in this experiment is $N=3$.

\paragraph{The convergence rate of our causal discovery method correlates with $\hat{\lambda}$.}
% \molko{NN opt not introduced}
%\mg{I  moved it after Synthetic data}
For each dataset we evaluate causal discovery on subsets of varying sizes. The size of the smallest subset required  for convergence is recorded and compared to value of $\hat{\lambda}$ of the dataset. The results are presented in Figure~\ref{fig:converge}. The experiment demonstrates that datasets with higher $\hat{\lambda}$ require on average fewer samples for successful recovery. The datasets, where convergence happened for 20 samples have $\hat{\lambda}$ of around $0.2$ (with a significant variance), while for datasets that require $8000$ samples the $\hat{\lambda}$ is exculsively under $0.05$. \piotrm{17.02 IMPORTANT: more comprehensive experiments?}
%\mg{Add a description of a graph}

%\molko{Case study whould suggest we are investigatin very specific dataset. Maybe we should only write about second experiment in this section and move the plot with SHD to section \ref{sec:nn_opt}? Otherwise we should change the name of this section.}
% In Section~\ref{sec:nn_opt}, we demonstrated that $\hat{\lambda}$ can effectively estimate the amount of data required to recover the true graph. 
%However, it is also important to quantify structural errors when recovery is incomplete, as this helps assess model reliability and  compare methods. Additionally, understanding how these errors decrease with more data provides insight into the efficiency of the learning process. \molko{12.02 This motivation is not convincing. Actually, do we need to motivate additional experiments?} \molko{I tired improving the intro with Chat, old version is in the comment below}
% In Section~\ref{sec:nn_opt} we showed that the $\hat{\lambda}$ can serve as a good proxy\molko{proxy is informal} for quantifying how much data is needed to uncover the true graph. We would like to know also how big are the errors in terms of structural metric when we do not uncover the true graph and how fast improvemnt on average of this metric can we observe.

\paragraph{Accuracy of causal discovery improves with data size.} Apart from convergence, we want to quantify structural errors when recovery is incomplete, to assess model reliability and  compare methods. Understanding how these errors decrease with more data provides insight into the efficiency of the learning process.
The results are presented in Figure~\ref{fig:naive_quantile}. For very small datasets we observe a relatively big $\eshdcpdag$ of 4, which rapidly improves with sample size.  As the sample size grows, the structure discovery accuracy stabilizes. For sample sizes of 2,500 and 8,000, the average value of $\eshdcpdag$ is just below 2. In the dataset used for this experiment, the average number of edges in CPDAG is around $8.4$, meaning that on average almost 25\% of the edges are predicted incorrectly.

As previously shown, we observe that datasets vary in difficulty. While average $\eshdcpdag$ remains distant from 0, the number of structures that were possible to identify using observational sample of each size increases. For $80$ samples there are $6$ graphs that do converge, for $250$ it is $26$ graphs. The rate of improvement slows down and between $2500$ and $8000$ it is respectively $34$ and $35$ graphs. 

% As expected, as the sample size increases, the number of graphs for which training successfully converges (i.e., recovers the true graph) also increases. We investigate the relationship between the minimum sample size required for convergence and the $\hat{\lambda}$. The results, presented in Fig.~\ref{fig:converge}, indicate that datasets with higher $\hat{\lambda}$ require fewer samples to converge. 
% However, there is considerable variance among distributions with similar $\lambda$, likely due to errors in $\lambda$ approximation and other distributional characteristics.

\begin{table*}[tbp]
    % \vspace{-1cm} % Adjust this as needed
    \small
    \centering
    \setlength{\tabcolsep}{4pt}
    \begin{tabular}{l|ll|ll|ll}
        & \multicolumn{2}{c|}{$\ERfive$} & \multicolumn{2}{c|}{$\ERten$} & \multicolumn{2}{c}{$\ERthirty$} \\
        \midrule
        Method & $\eshdcpdag$ & $\fscorecpdag$ & $\eshdcpdag$  & $\fscorecpdag$ & $\eshdcpdag$ & $\fscorecpdag$ \\
        \midrule
        DCDI &   5.7 \tiny{(3.7, 8.1)} & 0.60 \tiny{(0.46, 0.74)} & \textbf{16.9 \tiny{(15.7, 18.1)}} & 0.52 \tiny{(0.50, 0.56)} & \textbf{45.9 \tiny{(42.0, 49.9)}} & \textbf{0.73 \tiny{(0.69, 0.77)}} \\
        BayesDAG & 3.9 \tiny{(3.6, 4.3)} & 0.78 \tiny{(0.77, 0.81)} & 18.3 \tiny{(16.9, 19.8)} & \textbf{0.56 \tiny{(0.54, 0.59)}} & 51.7 \tiny{(48.2, 55.9)} & 0.59 \tiny{(0.57, 0.61)} \\
        DiBS & \textbf{2.6 \tiny{(1.7, 3.7)}} & \textbf{0.85 \tiny{(0.80, 0.90)}}& \textbf{16.9 \tiny{(14.2, 201)}} & \textbf{0.61 \tiny{(0.57, 0.68)}} & 68.0 \tiny{(65.3, 70.9)} & 0.23 \tiny{(0.22, 0.24)} \\
        SDCD & 5.4 \tiny{(3.8, 6.7)} & 0.60 \tiny{(0.35, 0.69} & 20.9 \tiny{(19.5, 22.2)} & 0.54 \tiny{(0.46, .62)} & 62.8 \tiny{(58.8, 67.7)} & 0.55 \tiny{(0.53, 0.58)} \\
        % DCDI &   5.70 \tiny{(3.70, 8.10)} & 0.60 \tiny{(0.46, 0.74)} & \textbf{16.92} \tiny{(15.66, 18.12)} & 0.52 \tiny{(0.50, 0.56)} & \textbf{45.87} \tiny{(41.99, 49.94)} & \textbf{0.73} \tiny{(0.69, 0.77)} \\
        % BayesDAG & \textbf{3.90 \tiny{(3.63, 4.33)}} & \textbf{0.78 	\tiny{(0.77, 0.81)}} & 18.26 \tiny{(16.89, 19.76)} & \textbf{0.56} \tiny{(0.54, 0.59)} & 51.72 \tiny{(48.24, 55.89)} & 0.59 \tiny{(0.57, 0.61)} \\
        % DiBS & & & 21.28 \tiny{(20.13, 22.47)} & 0.50 \tiny{(0.49, 0.52)} & 68.01 \tiny{(65.28, 70.85)} & 0.23 \tiny{(0.22, 0.24)} \\
        % SDCD & 5.40 \tiny{(3.80, 6.70)} & 0.60 \tiny{(0.35, 0.69} & 20.87 \tiny{(19.51, 22.24)} & 0.54 \tiny{(0.46, .62)} & 62.83 \tiny{(58.80, 67.74)} & 0.55 \tiny{(0.53, 0.58)} \\
        
    \end{tabular}
    \caption{Comparison of $\eshdcpdag$ and $\fscorecpdag$ for different methods on $\ERten$ (left) and $\ERthirty$ (right) dataset. The numbers in the subscripts correspond to 95\% confidence intervals. The statistics were computed based on 30 graphs.}
    \label{tab:combined_results}
\end{table*}

\subsection{Case study} \label{sec:case_study}

\begin{figure}[htbp]
    \centering
    \includegraphics[width=0.95\linewidth]{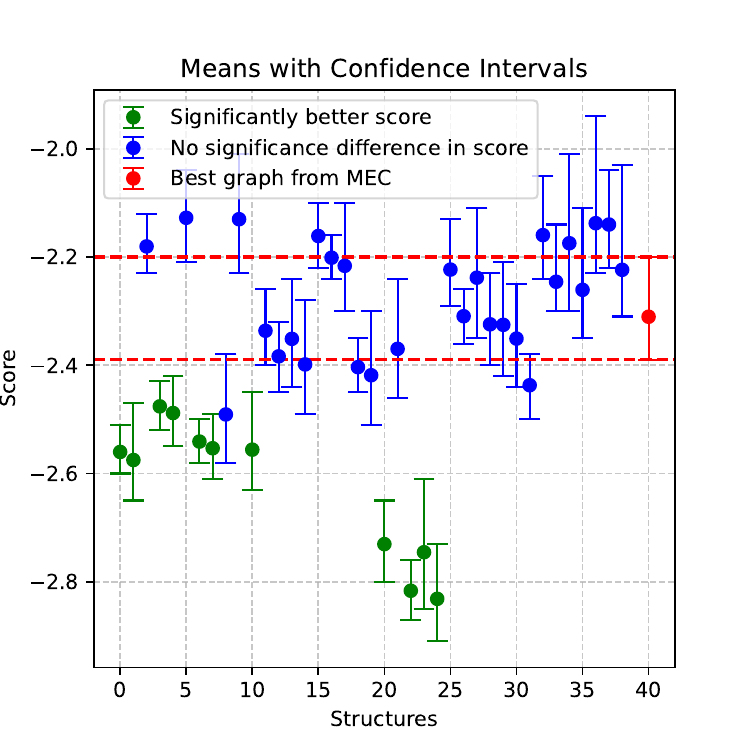}
    \caption{Exemplary results of score evaluation using our robust neural network based approximation approach. In red --- score of the target structure, in green --- scores of structures with statistically significantly better scores, in blue --- scores of structures with comparable scores. The error bars signify 95\% confidence intervals. Note that considerable number of structures has significantly better (lower) score that the target structure.}
    \label{fig:exp1}
\end{figure}

To further illustrate the challenges that are encountered during causal discovery with neural network we conducted a qualitative assessment. 
We sampled 10 datasets from SCMs constructed according to description above and use bootstrap of size $N = 29$. This allowed to compute 95\% bootstrap confidence intervals
of score for each DAG from $\mathcal{G}_5$ in each dataset, see Figure~\ref{fig:exp1} for exemplary results of the evaluation. 

In 6 out of 10 datasets the score of the target structure\footnote{We call \textit{the target structure} the structure from Markov Equivalence Class that achieved the highest average score.} overlaps or is worse than other graphs from outside of MEC.

In the extreme case there are 398 structures with overlapping confidence intervals and 11 structures with scores significantly better than the score of the target structure. 
% \molko{12.02 It seems that it is the first place where we use the argument that our method has small approximation error. Maybe we can move the discussion about approximation error here?}
Note that this evaluation is conducted in the class of graphs with only 5 nodes and the examples were sampled according to standard benchmarking practice and not prepared adversarially.
We argue that the existence of significant error in the score approximation indicates estimation errors due to finite sample dataset. We observe that the same training procedure applied to datasets of smaller size gives worse results (see Table~\ref{tab:better}), supporting the claim.

% It is highly unprobable that method will reach SHD=0 in any reasonable sample size. These
% findings demonstrate the methods’ consistent inability to
% identify correct structures. \molko{12.02 This comment is missplaced. Can we remove it entirely? We can metntion something similar in the conclusions.}

\begin{table}[H]
%     \centering
%     \begin{tabular}{c|c|c|c}
%         {Number of samples} & Mean & Median & Max\\
%         \midrule
% $800$ & 45.5 & 3.0 & 398 \\
% $2500$ & 7.1 & 0.0 & 36\\
% $8000$ & 1.3 & 0.0 & 11 \\
%         \midrule
%     \end{tabular}
%      \caption{Statistics of graphs with better }
%           \label{tab:better}
    \centering
    \begin{tabular}{c|c|c}
        {Number of samples} & Mean & Max\\
        \midrule
$800$ & 45.5 & 398 \\
$2500$ & 7.1  & 36\\
$8000$ & 1.3  & 11 \\
        \midrule
    \end{tabular}
     \caption{Statistics of structures with significantly better score than than the \textit{target structure}. Data aggregated from 10 randomly sampled SCMs.}
          \label{tab:better}
\end{table}

\subsection{Conclusions}
In this section, we provided a detailed experimental analysis of the relationship between the $\lambda$-strong faithfulness property of non-linear datasets and the performance of causal discovery with neural function approximators. Our results indicate that the $\lambda$-strong faithfulness property constrains causal discovery from finite data, limiting both convergence rate and accuracy.
Furthermore, we demonstrate in a controlled setting 
% the impact of limited data on likelihood-based score estimation and demonstrate
that data scarcity introduces significant errors in score assignment. 
Additionally, we show that the proportion of $\lambda$-strong faithful distributions in ER graphs with nonlinear functions rapidly decreases as graph density and size increase. 
Those findings, along with the theoretical results of \citet{geometry_of_faithfulness}, suggests that current neural causal discovery methods may face fundamental limitations. 
\section{Evaluation of Neural Causal Discovery Methods}  \label{sec:benchmark}

%After showing that the neural networks come across several challenges and connecting this phenomenon to $\lambda$ - strong faithfulness in Sections~\ref{sec:nn_opt} and~\ref{sec:difficulty}, in this section, we investigate how this violation affects neural causal discovery methods performance. To this end we craft a systematic benchmarking protocol to ensure valid and fair results.  

\mg{refurbished intro}
After linking the challenges in neural causal discovery to $\lambda$ - strong faithfulness in Sections~\ref{sec:nn_opt} and~\ref{sec:difficulty}, in this section, we investigate how this violation affects neural causal discovery methods performance in terms of structural metrics.\mg{performance, but specified} Additionally since these methods were constructed with scalability in mind, the experiments can be performed on graph bigger that in Section~\ref{sec:main}.

% To prove that the problem indeed manifests in contemporary neural causal discovery methods we propose a practical metric that quantifies how distribution adheres to the strong faithfulness property of the distribution given finite dataset and its underlying structure. We show that it highly correlates with the performance of selected neural discovery methods on our systematic benchmark.

\subsection{The systematic benchmarking protocol}\label{sec:benchmark:experimental_setup}
%Our findings would be impossible to describe if it was not for our unified benchmarking effort.
We evaluate methods DiBS, DCDI, BayesDAG, and SDCD introduced in Section~\ref{sec:background} on identical datasets, tune hyperparameters consistently, and use a common functional approximation.

\paragraph{Dataset generation}
We sample three types of graphs from the Erdős-Rényi (ER) distribution~\citep{er_graphs} as described in~\ref{sec:difficulty}: one with 5 nodes and the expected degree of 1, another with 10 nodes and the expected degree of 2, and the third with 30 nodes and the expected degree of 2. These datasets are referred to as $\ERfive$, $\ERten$, and $\ERthirty$, respectively. These parameter choices align with commonly studied medium-sized graphs in causal discovery research~\citep{dcdi, sdcd}. 
%Data generation follows the SCM formalism introduced in Section~\ref{par:scm}, with functional relationships modeled by two-layer neural networks (hidden dimension 8, ReLU activation) and additive Gaussian noise. The noise has zero mean, and its variance is sampled independently for each node. This setup is known to be challenging~\citep{deependtoendcausalinference, sdcd}. 

\newcommand{\myspacefour}[0]{-0.1em}
\vspace{\myspacefour}

\paragraph{Hyperparameter tuning}\label{sec:hyperparameters}
To ensure a fair comparison across all methods, we perform systematic hyperparameter tuning, selecting the best-performing parameters for each method
%\molko{model? maybe approach/method?}. 
We employ a grid search approach based on the parameter ranges suggested by the original authors. This process optimizes key variables such as regularization coefficients, sparsity controls, and kernel configurations. Details can be found in Appendix~\ref{appendix:hyperparameters_grid}.
\vspace{\myspacefour}

\paragraph{Functional approximators}
We standardize the choice of functional approximators across all experiments, using a two-layer MLP with a hidden dimension of 4 to model each functional dependence $f$ (see Section~\ref{sec:background}). This model size is consistent with previous work~\citep{dcdi, sdcd} and has proven to perform well across all the benchmarked methods, as discussed in Appendix~\ref{sec:benchmark:models}. Additionally, we use trainable variance to allow the model to adapt to varying noise levels, in line with our dataset generation setup. 
% \alicja{This section makes me feel suspicious, especially, since both of the works you cite that use this architecture are earlier classified as the same type of approach - non-bayesian. And the the use of the word most also makes me feel a little bit uneasy. Especially, since you want to say that the performance does not scale with the data, and those networks are quite tiny in comparison to networks normally used in ML, which would cause lack of scaling with the data.}
\vspace{\myspacefour}
\paragraph{Structure evaluation} We evaluate graph discovery within the MEC using $\eshdcpdag$ and $\fscorecpdag$, where $\eshdcpdag = 0$ and $\fscorecpdag = 1$ when the predicted graph is in the same MEC as the ground truth. 
For Bayesian methods, we computed the expected value by sampling from the posterior distribution; for non-Bayesian methods, we use a single graph.
%For Bayesian methods, we compute the average by sampling 100 graphs from the posterior; for non-Bayesian methods, we use a single graph. 

% \molko{$\rightarrow$ For Bayesian methods we estimate expected value using Monte Carlo, for non-Bayesian methods, we use a single graph.} \mg{I think we should not use MOnteCarlo, as it will add more unnesesarly complexity, I prpopose something else.}

% \molko{With dibs we use  30 graphs as this is the default number of particles, lets put this info in the appendix.} \mg{Now there are no numbers, but propably still worth to put it into the appendix.}

The Structural Hamming Distance (SHD)~\citep{shd} counts edge insertions, deletions, and reversals needed to match the predicted graph to the true graph.
% For example, SHD = 10 for a graph with $n=10$ and average degree $d=2$ implies 10 structural modifications are needed.
%Lower SHD indicates better recovery of the true causal structure.
We define \textbf{Expected SHD between CPDAGs} as:

\begin{multline} \eshdcpdag(\mathcal{G}, \sG) = \\
= \mathbb{E}_{\mathcal{G}^* \sim \sG} [\text{SHD}(\text{CPDAG}(\mathcal{G}), \text{CPDAG}(\mathcal{G}^*))], \end{multline}

where $\sG$ is the resulting distribution of graphs, $\mathcal{{G}^*}$ is a graph sampled from $\sG$ and $\mathcal{{G}}$ is the ground true graph.
% \paragraph{Expected F1-Score between CPDAGs}
% The F1-Score measures the harmonic mean of precision and recall for edge predictions, where precision reflects the fraction of correctly predicted edges among all predicted edges, and recall reflects the fraction of correctly predicted edges among the true edges.
The F1-Score measures the harmonic mean of precision and recall for edge predictions.
We compute the \textbf{Expected F1-Score between the CPDAGs} as follows:

\begin{multline} \fscorecpdag(\mathcal{G}, \sG) = \\ = \mathbb{E}_{\mathcal{G}^* \sim \sG} [\text{F1-Score}(\text{CPDAG}(\mathcal{G}), \text{CPDAG}(\mathcal{G}^*))]. \end{multline}
For more details and justification on the selection of metrics please refer to Appendix~\ref{app:metrics_justification}.

% \subsection{Results}
% \mikolajm{I find these introductions to sections quite useless. You are not presenting the broad overview of the structure of the paper, you are telling the reader what will be presented on the next single page. It seems to me that these introductory paragraphs are more space fillers than anything else}

\paragraph{Methods comparison}
% \molko{Change to subsection?}
\label{sec:benchmark:main}

Table~\ref{tab:combined_results}, summarizes the benchmark results of neural-based causal discovery methods on graphs from \ERfive{}, $\ERten$, and $\ERthirty$ classes. We tune hyperparameters to optimize the $\eshdcpdag$ metric. For all classes of graphs, metrics were computed based on 30 graphs. \molko{is this still true?}

The results show that DiBS is particularly effective for smaller graphs (\ERfive{} and \ERten{}), while DCDI is able to achieve the best results for moderate-size graphs (\ERten{} and \ERthirty{}). The ranking of the methods changes with the size of the graphs but SDCD consistently exhibits the worst performance in terms of $\eshdcpdag$.
% The results on larger graphs show that DCDI, the earliest approach here, achieves the best $\eshdcpdag$ score and the best or comparable $\fscorecpdag$ to all other methods on both graphs. 
% Moreover, the performance gap widens with the size of the model with the second best, the BayesDAG method, being 8\% worse than DCDI on \ERten{} and 13\% worse on \ERthirty{}.  

Nevertheless, the structural metrics of all the methods remains unsatisfactory with all methods predicting more than half of the edges incorrectly.
% Additional results, on $\ERfive$ dataset, are in Appendix~\ref{appendix:p5e5}.
% \reb{For \ERfive{} trends in the presented results are slightly different, with DiBS performing the best in terms of $\eshdcpdag$ and $\fscorecpdag$, and having the smallest confidence interval. Nevertheless, all methods predict more than half of the ground truth edges wrong.}

% \molko{The results paint a clear picture. DCDI, the earliest approach here, achieves the best $\eshdcpdag$ score and the best or comparable $\fscorecpdag$ to all other methods on both graphs. Moreover, the performance gap widens with the size of the model with second best, BayesDAG method, being 8\% worse than DCDI on \ERten and 13\% worse on \ERthirty.  Nevertheless, the performance of the methods remains unsatisfactory with all methods predicting more than half of the edges incorrectly.}

% For the $\ERten$ dataset, results across methods are fairly consistent, with DCDI, one of the earliest approaches, achieving the best $\eshdcpdag$ score, though its $\fscorecpdag$ was slightly lower but still comparable. In contrast, performance diverged on the $\ERthirty$ dataset, where scalability improvements did not result in higher causal discovery accuracy, with all methods predicting more than half of the edges incorrectly.

\subsection{
$\hat{\lambda}$ and NCD method performance}

% \molko{Sketch:
% \begin{itemize}
%     \item Recall lambda-fiath property. Shand and Sprites write that lambda can serve as notion of robustness against apparent faithfulness violations in finite data.
%     \item It is impossible to directly estimate lambda for finite dataset due to irreducible approximation errors.
%     \item So instead of selecting a lambda-threshold, we quantify predictive is correlation on  d-sepearation for any threshold.
%     \item Observe that for fixed data size. Datasets where correlation is stronger will likely obtain higher values of this metric since the probability that the correlation value will be greater is higher.
% \end{itemize}
% }

\begin{figure*}[htbp]
\begin{subfigure}[t]{0.5\textwidth}
    \centering
    \includegraphics[width=0.8\textwidth]{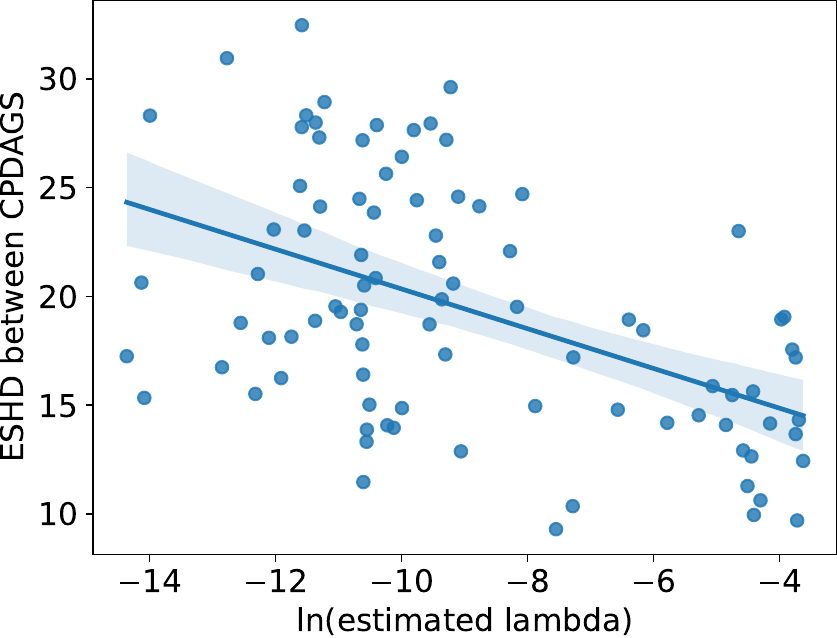}
    \caption{}
    \label{fig:faith_perf_corr}
\end{subfigure}
~
\begin{subfigure}[t]{0.5\textwidth}
    \centering
    \includegraphics[width=0.9\textwidth]{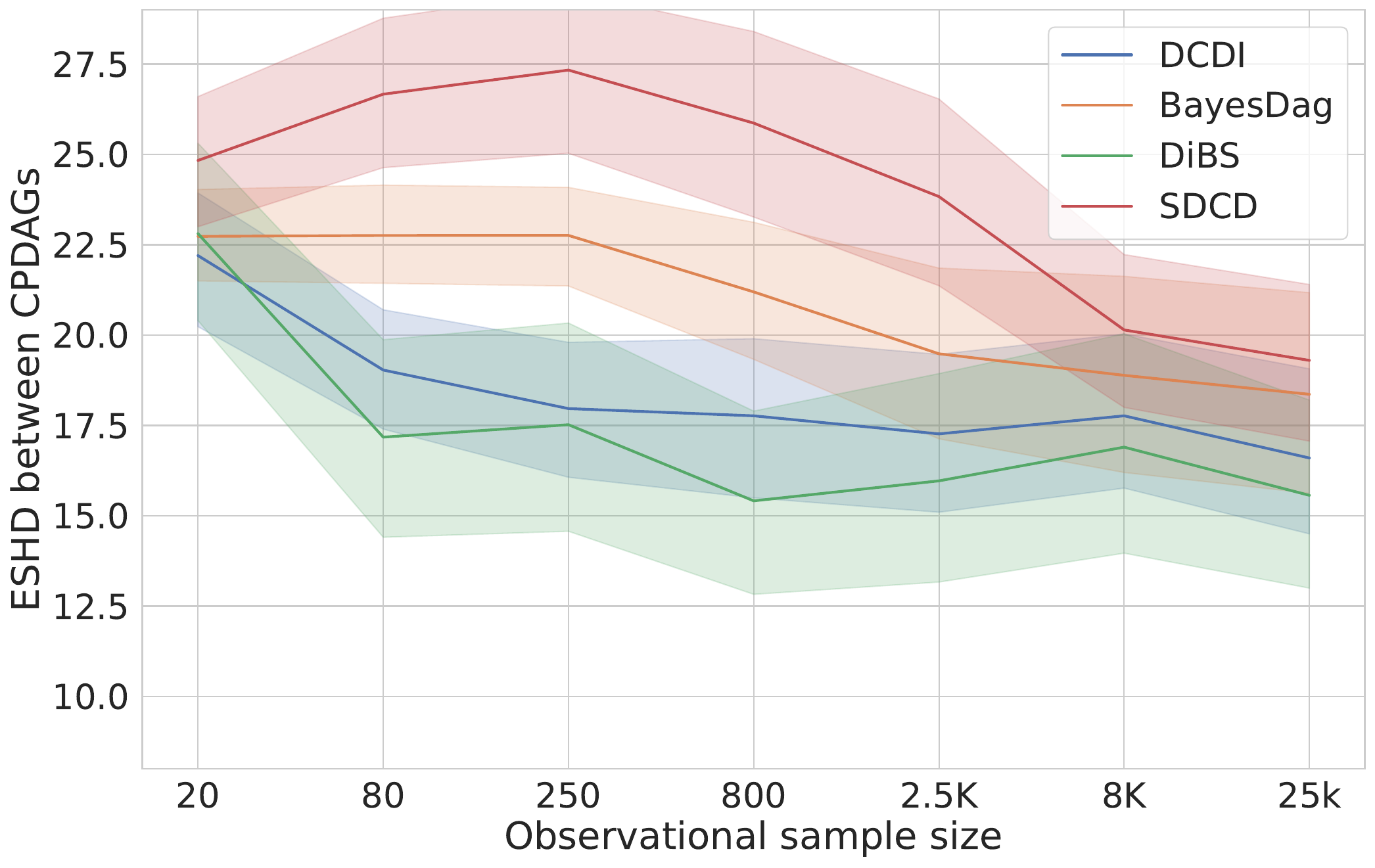}
    \caption{}
    \label{fig:p10e20_comparision_sample_size}
\end{subfigure}
\caption{(a) Linear regression fit between the average performance of neural causal discovery methods and $ln(\hat{\lambda}$ measure. The p-value for spearman rank correlation between $\hat{\lambda}$ and  $\eshdcpdag$ and  is 4e-6, signifying anti-monotonic correlation. b) Performance of benchmarked methods in terms of $\eshdcpdag$ with resect to dataset size for $\ERten$ graphs, averaged over 30 samples. }
\vspace{-1em}
\end{figure*}

To investigate the tie between $\hat{\lambda}$ metric and the difficulty of neural causal discovery, we generate 30 graphs from the $\ERten$ class, introduced in Section~\ref{sec:benchmark:experimental_setup}. Based on each graph, we define three different SCMs, resulting in 90 distinct distributions. From each one we generate 8,000 observational samples.
We then evaluate the $\hat{\lambda}$ of each dataset and compute the performance of the selected neural-based causal discovery methods.

% \paragraph{Results}
% \mikolajm{Since Spearman's rank correlation does not measure linear agreement but only monotonic agreement, there is no reason to limit ourselves to a linear fit on the graph. We can perform any nonlinear fit. The result of -0.58 indicates a very strong anti-monotonic relationship, which is entirely sufficient to confirm the main conclusion: greater DeFaith leads to lower SHD (and this relationship does not have to be linear). I would suggest improving the figure and replacing the linear regression (LR) with a nonlinear fit.}\molko{Thanks w could not do it due to lack of time - rebutal}
In Figure~\ref{fig:faith_perf_corr} we present the relationship between average performance of all methods and the logarithm of $\hat{\lambda}$ (for readibility) for all 90 distributions in the dataset. The performance is better (lower SHD) for distributions with higher $\hat{\lambda}$. The Spearman's rank correlation coefficient is $\rho=-0.46$ and p-value for sperman correlation test is $4e-6$.
This result proves the strong anti-monotonicity between $\hat{\lambda}$ and average methods' performance.

\subsection{Lack of scalability}

\label{sec:benchmark:samples}
%We investigate the impact that the sample size has on method's performance.
We want to investigate the impact that the sample size has on a causal discovery methods. 
We expect it to behave similar to our findings from Section~\ref{sec:nn_opt} as we hypothesize that the lack of performance is due to relation between data and approximators. We would like to note that the NCD methods use one neural network to approximate the distribution under changing set of parents, which is harder task than the one in Section~\ref{sec:nn_opt}.
We evaluate the benchmarked methods 
%\molko{which methods?} 
on datasets with varying number of observational samples, ranging from 20 to 8,000 observations.

The results, presented in Figure~\ref{fig:p10e20_comparision_sample_size}, reveal no consistent pattern of improvement in the $\eshdcpdag$ metric as observational sample size increases, despite extensive hyperparameter tuning (as described in Section~\ref{sec:hyperparameters}). For example, DiBS shows the best performance on larger datasets, but its improvements plateau after around 800 samples. Similarly, BayesDAG shows only marginal improvements with larger sample sizes and is unable to outperform DiBS. DCDI improves up to 250 samples and then maintains consistent performance regardless of the sample size, similar to DiBS. Interestingly, SDCD's performance is poor on datasets with small number of observations but begins to improve once sample sizes exceed 250, though is unable to reach DCDI's performance, for larger sample sizes the rate of improvement decreases.

Further analysis of the effect of sample size on smaller graphs \ERfive is presented in Figure~\ref{fig:p5e5_samples} in Appendix~\ref{p5e5_samples}. Overall, the results on smaller graphs align with the trends observed on larger graphs. Specifically, while some methods improve with increasing sample size, others show inconsistent or even degraded performance.

Even for the bigger sample sizes the results remains unsatisfactory. This confirms our observations, made in the previous sections about datasets difficulty and the amount of samples required for convergence. We conclude, that the amount of data needed to recover a graph from a true MEC with those methods is out of scope for most applications.

\section{Limitations and Future work}
\molko{31.01 Section freezed}
\begin{itemize}
\item Work of \citet{enco} suggests that interventional data can replace the need for faithfulness assumption. A valuable extension of our research would be to evaluate the performance of the benchmarked methods on interventional datasets to understand their limitations and potential improvements in this context.

\item Our work provides experimental evidence for the challenging nature of causal discovery from non-linear data, especially when using neural networks. 
It would be beneficial for the community to establish theoretical bounds on the best possible performance in such cases. 

\item Neural networks are well known approximators, with well established position in Machine Learning community. It will be however interesting to investigate other classes of aproximators, that could possibly have a better suited characteristic for the task of causal discovery.

\item In this work we provided results for the Erodos-Renyi class of graphs. The results could be computed for more classes such as for example Scale Free graphs.

\item Our conclusions are based on characteristic of distributions associated with Bayesian Networks provided by \citet{geometry_of_faithfulness}. The result says that for any fixed $\lambda$, $\lambda$-strong faithful distributions vanish exponentially with the size of the graph. It may hold, though is highly unprobable, that real-world distributions adhere to $\lambda$-strong faithfulness despite large sizes of the graph. Further investigation is required.
\end{itemize}
% \asia{only a few ER datasets, no real-world datasets, we hope that these results can be theoritically proven}

\section{Discussion}
In this section, we provided a detailed experimental analysis of the relationship between the $\lambda$-strong faithfulness property of non-linear datasets and the performance of causal discovery with neural function approximators. Our results indicate that the $\lambda$-strong faithfulness property constrains causal discovery from finite data, limiting both convergence rate and accuracy.
Furthermore, we show that the proportion of $\lambda$-strong faithful distributions in ER graphs with nonlinear functions rapidly decreases as graph density and size increase. 
We postulate that improvements in the causal discovery efficiency and approximation accuracy have saturated and now the domain faces fundamental limitations.

To the best of our knowledge, we are the first to demonstrate connection between $\lambda$-strong faithfulness and performance of score-based methods, previously discussed only for independence testing approach \citep{zhang2003faithfulness}. 
% We show that performance of the neural causal discovery methods is limited by the $\lambda$-strong faithfulness property of the datasets. Additionally we observe that for a fixed $\lambda$, $\lambda$-faithful distributions vanish rapidly with the size of the graph (as show by \citet{geometry_of_faithfulness}. 
% % This suggest that causal discovery on large graphs using neural-based approach is fundamentally limited. 
% % We briefly discuss alternative approaches that could potentially offer a solution to the problem.
% We postulate that improvements in the causal discovery efficiency and approximation accuracy have saturated and now the domain faces fundamental limitations.

Our results signify the need to evaluate causal discovery methods on real world data, which may exhibit different characteristics than uniformly initialized neural networks. 
A very interesting approach was recently proposed was recently proposed by~\citep{gamella2025causal} where a physical simulation of a machine was used to create causal data. It brings the benefit of knowing the true graph while preserving the characteristics of data generated by real life physical systems.  

While many causal discovery methods rely on the faithfulness assumption, alternative conditions or relaxation have been proposed.
In the context of linear structural causal models (SCMs), \citet{van2013} demonstrated that a sparsity-based assumption can effectively reveal the underlying causal structure.
More recently, \citet{Ng2021Reliable} suggested another causal discovery method, based on relaxed faithfulness assumption that requires less independence tests to be fulfilled.
% (Sparsest markov representation plays a role?) what are the limitations, what type of data it assumes?
An idea, that have not yet get much attention is relaxing the task of causal discovery and discovering only some independencies relations in data, like in~\citep{amini2022non}.

Overall, we do believe that addressing the challenges pointed out in this work, will help to navigate towards more robust methods, that will deliver satisfactory results in real world scenarios.

\molko{Consider writting something about amortized approaches. Are they a viable option to solve problems pointed by our work?}

% \paragraph{Amortized methods}

% \paragraph{Structure discovery with less restrictive assumptions}

% \paragraph{Conservative approaches to CD}

\section*{Impact Statement}
This paper presents work whose goal is to advance the field of Machine Learning. There are many potential societal consequences  of our work, none which we feel must be specifically highlighted here.

\bibliography{iclr2025_conference}
\bibliographystyle{icml2025}

\newpage
\appendix
\onecolumn

\section{Additional background information}
% \reb{New name of the section}
\subsection{$d$-separation}\label{appendix:d_separation}

% \mikolajm{The introduction of the faithfulness assumption happens "out of the blue", I don't think that ICLR reviewers will be familiar with this wording, not to mention that the definition below uses the undefined notion of $d$-separation. My suggestion is to expand the paragraph by first defining $d$-separation, and then explaining the dual nature of Markov assumption and faithfulness assumption}

% \paragraph{Faithfulness Assumption}
% Before introducing the faithfulness assumption, let us first define the notion of $d$-separation.
Two nodes $A$ and $B$ in a DAG are said to be \textbf{$d$-separated} by a set of nodes $Z$ if all paths between $A$ and $B$ are blocked when conditioning on $Z$. A path is considered blocked under the following conditions:
\begin{itemize}
    \item If a path includes a non-collider node (a node where arrows do not converge, i.e., a chain or fork), conditioning on that node blocks the path. For example, if $A \rightarrow C \rightarrow B$, or $A \leftarrow C \rightarrow B$, conditioning on $C$ makes $A$ and $B$ independent.
    % \item If a path includes a collider (a node where arrows diverge), the path is only blocked if neither the collider nor any of its descendants are conditioned on. For instance, in the path $A \rightarrow C \leftarrow B$, conditioning on $C$ or its descendants would activate the path, making $A$ and $B$ dependent.
    % \item If there are multiple paths connecting $A$ and $B$, all must be blocked for $A$ and $B$ to be considered d-separated.

    \item If the path includes a collider (a node where arrows converge, i.e., \(A \rightarrow C \leftarrow B\)), the path is blocked unless either the collider itself or one of its descendants is conditioned on. For instance, in the path \(A \rightarrow C \leftarrow B\), conditioning on \(C\) or its descendants would unblock the path, making \(A\) and \(B\) dependent.

    \item  If there are multiple paths connecting \(A\) and \(B\), all paths must be blocked for \(A\) and \(B\) to be considered $d$-separated. Even if one path remains unblocked, \(A\) and \(B\) are d-connected, meaning they are dependent.

\end{itemize}

In causal discovery, we are interested in making statements about the relationship between the causal graph and the data distribution. Given a causal graph $G$ and the data distribution $P$, the \textbf{Markov assumption} states that if variables $A$ and $B$ are $d$-separated in the graph $G$ by some conditioning set $C$, then $A$ and $B$ are conditionally independent in the distribution $P$ when conditioned on the same conditioning set $C$. Formally, this can be written as:

\begin{equation}
A \independent_G B | C \Rightarrow A \independent_P B | C
\end{equation}

\subsection{Example of faithfulness violation} \label{appendix:faith_example}

In this subsection we will illustrate a faithfulness violation for a simple $3$ nodes structural causal model with linear functions and additive Gaussian noise. Such a setup is aimed at showing example of faithfulness violation while maintaining simplicity.
The example and graphics is from~\citep{geometry_of_faithfulness}.

\begin{figure}[ht]
    \centering
    \includegraphics[width=0.2\linewidth]{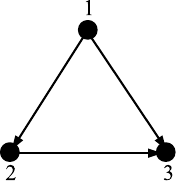}
    \caption{Simple 3 nodes graph $G$.}
    \label{fig:3node}
\end{figure}

\newcommand{\cov}{\operatorname{cov}}
\newcommand{\corr}{\operatorname{corr}}
First lets define a structural causal model on a graph $G$ shown in graph~\ref{fig:3node}.

\begin{eqnarray*}
X_1 &=& \varepsilon_1,
\\
X_2 &=& a_{12} X_1 + \varepsilon_2,
\\
X_3 &=& a_{13} X_1 + a_{23}
X_2 + \varepsilon_3,
\\
(\varepsilon_1, \varepsilon_2, \varepsilon_3) &\sim& \mathcal{N}(0,I),
\end{eqnarray*}

Since data is linear we can use covariance to measure dependency of variables. Using defined structural causal model, we can write:

\begin{eqnarray}
\label{eqq1}
\cov(X_1,X_2) & = & a_{12},
\\
\label{eqq2}
\cov(X_1, X_3) & = & a_{13} +
a_{12}a_{23},
\\
\label{eqq3}
\cov(X_2, X_3) & = & a_{12}^2
a_{23} + a_{12}a_{13}+ a_{23},
\\
%&&\nonumber\\
\label{eqq4}
\cov(X_1, X_2 \mid X_3) & = &
a_{13} a_{23}-a_{12},
\\
\label{eqq5}
\cov(X_1, X_3 \mid X_2) & = & -
a_{13},
\\
\label{eqq6}
\cov(X_2, X_3 \mid X_1) & = & -
a_{23}.
\end{eqnarray}

If we define $a_{13}, a_{23}, a_{12}$ in such a way that:
$$a_{13} * a_{23} -  a_{1,2} = 0$$
then we get a situation where:
nodes $1$ and $2$ are not d-separated given node $3$  in a graph $G$ and $X_1 \independent X_2|X_3$ which is a violation of faithfulness. 

\section{\submethod{} details} \label{app:nn_opt}
\paragraph{Details of experiments with \submethod{}}In order to test which architecture perform best, we conducted an experiment, training \submethod{} with different sizes of neural networks. The trained models were judged in terms of negative log likelihood and their performance on the task of causal discovery measured as $\eshdcpdag{}$.
For each tested architecture, we performed the search for the best regularization coefficient, the tested coefficients were: $[0.1, 0.3, 1.0]$. Among all models, the best results were consistently obtained for regularization coefficient = $
0.3$. The learning rate was set to $0.0003$.
The results of the experiments are shown in Table~\ref{tab:opt_models}. As we can see, the best , both in case of NLL and $\eshdcpdag{}$ was model with two layers and hidden dimension of size 8. Notably this is the same architecture, as was used to generate data.

\textbf{Selected hyperparameters:} Number of layers = $2$, hidden dimension = $8$, regularization coefficient = $0.3$.
% \textbf{Grid search}: Regularization coefficients tested: [0.01, 0.1, 0.3, 1.0]. Values of 0.3 consistently performed better across all data sizes. MLP architectures tested: neural networks with [1, 2, 3] layers and hidden dimension in:[4, 8, 16]. MLP with hidden size of 8 and 2 layers performed best in terms of NLL and $\eshdcpdag{}$. In terms of regulariation coefficient the threshold = $0.3$ performed best, independent from model. \textbf{Selected}: regularization coefficient = $0.3$, hidden dimension = $8$, number of layers = $2$. Learning rate = $0.0003$. The experiment were performed on 30 graphs.
\begin{table}[htbp]
    \centering
    \begin{tabular}{l|l|l}
        {Model architecture} & NLL & $\eshdcpdag$\\
        \midrule
$[4]$&\ 0.33\tiny({0.22}, {0.43})& 3.63\tiny({2.83}, {4.67}) \\
$[4, 4]$&\ 0.2\tiny({0.1}, {0.3}) & 3.15\tiny({2.0}, {4.65})\\
$[4, 4, 4]$&\ 0.23\tiny({0.14}, {0.34}) & 3.03\tiny({2.33}, {4.07}) \\
$[8]$&\ 0.18\tiny({0.06}, {0.29}) & 2.13\tiny({1.43}, {3.07}) \\
$[8, 8]$& \textbf{0.13}\tiny({0.02}, {0.24}) & \textbf{1.23}\tiny({0.77}, {1.87})\\
$[8, 8, 8]$&\ 0.22\tiny({0.12}, {0.32}) & 2.77\tiny({1.97}, {3.67})\\
$[16]$&\ 0.14\tiny({0.03}, {0.26}) & 1.77\tiny({1.1}, {2.73})\\
$[16, 16]$&\ 0.33\tiny({0.24}, {0.42}) & 2.4\tiny({1.0}, {4.32}) \\
$[16, 16, 16]$&\ 0.88\tiny({0.8}, {1.0}) & 4.0\tiny({3.07}, {4.97})\\

        \midrule

    \end{tabular}

     \caption{The performance of \submethod{} models with different architectures. The numbers in the subscripts, correspond to $0.95$ confidence intervals. The experiments were performed on 30 graphs.}
          \label{tab:opt_models}

\end{table}

\section{Errors in $\hat\lambda$ computation}
\label{sec:hat_lambda_errors}
To evaluate consisistency of $\hat\lambda$ computation. We conduct the following experiment. We generate 90 SCMs based on graphs from ER(5, 5) class and using NN-based functional relations (two layers of width 8 and ReLU activation). From each SCM we sample 10 datasets of size 10 000 samples. Then, we compute $\hat\lambda$  on each dataset. In Table~\ref{tab:hat_lambda_errors} we report average standard deviation of the $\hat\lambda$ values obtained for each SCM, as well as average standard deviation normalized by the average value. The results are splitted on various ranges of $\hat\lambda$. \molko{Conclusions?}

\begin{table}[H]
    \centering
    \begin{tabular}{c|ccc}
        $\hat\lambda$ range & \#SCMs & std & std / $\hat\lambda$ \\
        \midrule
0.1  - 1.0 & 16 & 0.019 & 0.112 \\
0.03 - 0.1 & 25 & 0.019 & 0.380 \\
0.01 - 0.03 & 27 & 0.009 & 0.481 \\ 
0    - 0.01 & 22 & 0.004 & 0.705 \\
\bottomrule
    \end{tabular}
    \caption{Consistency of $\hat\lambda$ approximation.}
    \label{tab:hat_lambda_errors}
\end{table}

\section{Details About Benchmark and Extensions}

\subsection{Dataset generation details} \label{appendix:dataset_generation}

The data is generated using a fully connected MLP with two hidden layers of 8 units each, initialized with random weights drawn from a uniform distribution and use the ReLU \citep{relu} activation function to introduce non-linearity. The neural network models the relationships between variables in the underlying DAG, where each node represents a variable and the edges capture dependencies between these variables. The input variables, which serve as the initial causes in the graph, are sampled from normal distributions. The noise added to the system is sampled from a Gaussian distribution $\mathcal{N}(0, 0.1^2)$, simulating uncertainty in the model. The dataset consists of 100,000 data points, and the data is rescaled to maintain consistency across samples.

\subsection{Model hyperparameters}\label{appendix:hyperparameters_grid}

We performed extensive hyperparameter tuning for all methods. In addition to the MLP architecture grids described in Appendix~\ref{sec:model_architecture}, the following hyperparameter grids were explored:

\paragraph{DCDI} \textbf{Grid search}: Regularization coefficients tested: [0.1, 0.3, 1, 2]. Values below 0.001 or above 5 led to poor performance.
\textbf{Selected}: Regularization coefficient = 1, learning rate = $0.001$, Augmented Lagrangian tolerance = $10^{-8}$.

\paragraph{DiBS} \textbf{Grid search}: Alpha linear: [0.01, 0.02, 0.05], kernel parameters: h latent: [0.5, 1.0, 2.0], h theta: [20.0, 50.0, 200.0], step size:[0.05, 0.03, 0.01, 0.005, 0.003].
\textbf{Selected}: Alpha linear = 0.02, h latent = 1.0, h theta = 50.0, \reb{step size = 0.03}.

\subsection{Testing models architecture}
\label{sec:benchmark:models}

\begin{figure*}[tbp]
    \centering
    \includegraphics[width=0.8\linewidth]{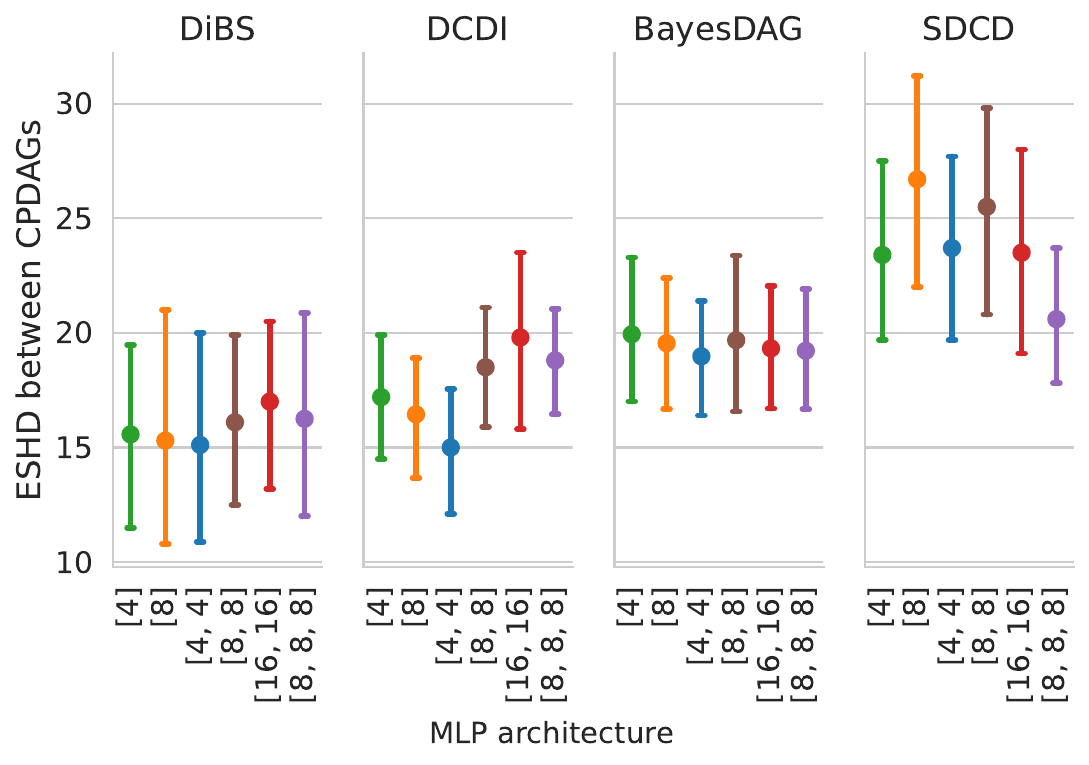}
    \caption{Comparison of $\eshdcpdag$ using different MLP architectures as functional approximator for $\ERten$ dataset and 800 observational  samples, averaged over 30 samples.}% \molko{@JW lest modify the sequence of models, lets do blue yellow, green, grown, red purple}}
    \label{fig:model_architecture_benchmark}
\end{figure*}

Finally, we investigate the impact of the neural model architecture, used as the functional approximator, on the performance of the benchmarked methods. Specifically, we assess how the capacity of different architectures influences the ability to uncover causal relationships from synthetic data. To provide a comprehensive evaluation, we explored architectures with 1, 2, and 3 layers, configured with 4, 8, and 16 hidden units.

Results, presented in Figure~\ref{fig:model_architecture_benchmark} show the comparison of $\eshdcpdag$ metric for the benchmarked architectures across all methods on dataset with 800 samples. We find that the choice of neural architecture has no significant impact on performance across methods. We conclude that any of the tested MLP architectures provides sufficient capacity to model the underlying distribution effectively. 
{Additionally for BayesDAG and SDCD we implemented layer normalization and residual connections. We investigated the impact of this changes in architectures and did not found any significant differences, see Figure~\ref{fig:model_options}. The details and additional experimental results are in Appendix \ref{sec:model_architecture}.

\paragraph{BayesDAG} \textbf{Grid search}: Scale noise: [0.1, 0.01], scale noise p: [0.1, 0.01, 1.0], lambda sparse: [50.0, 100.0, 300.0, 500.0].
\textbf{Selected}: Scale noise = $0.1$, scale noise p = $0.01$, lambda sparse = $500.0$.

\paragraph{SDCD} \textbf{Grid search}: Constraint modes: ["exp", "spectral radius", "matrix power"]. The $\eshdcpdag$ metric showed similar results across modes.
\textbf{Selected}: Spectral radius was chosen for faster computation, with a learning rate of $0.0003$.

For each of these method, all other parameters were retained from the original paper or code.
\subsection{Model architecture comparision within method}\label{sec:model_architecture}

\paragraph{DCDI}\label{model_architecture_comp_dcdi}

\begin{figure}[ht]
    \centering
    \includegraphics[width=1\linewidth]{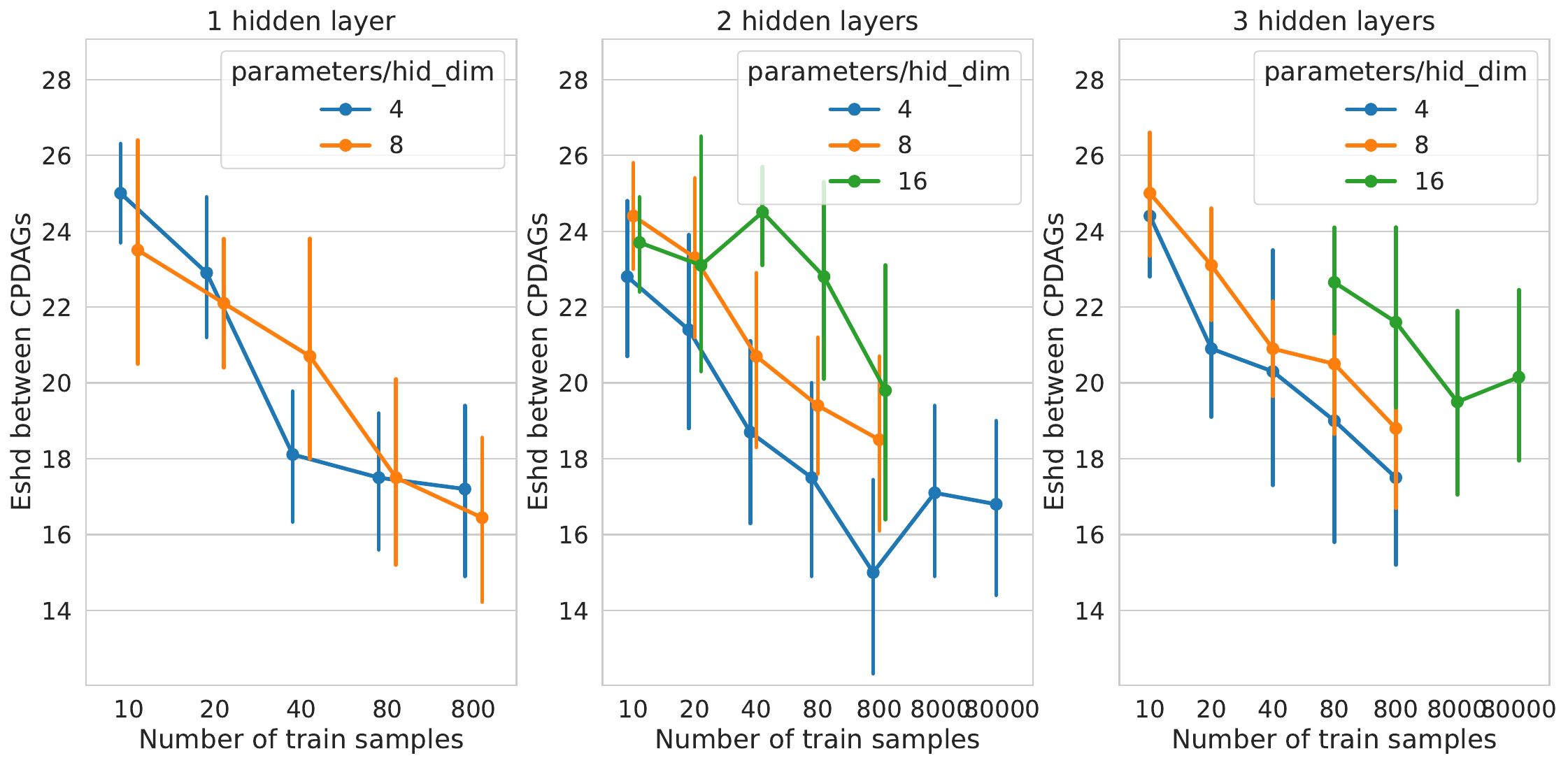}
    \caption{Comparison of the $\eshdcpdag$ of DCDI for datasets with different observational sample size. The result is based on 10 graphs.}
    \label{fig:dcdi_grid}
\end{figure}

In Figure \ref{fig:dcdi_grid}, we present the performance analysis of the DCDI across various neural network configurations.
Our results reveal that the optimal performance is generally achieved by a two-layer model with a hidden dimension of 4. Interestingly, we observe that more expressive models exhibit diminished performance relative to the smaller models.

\paragraph{DiBS}\label{model_architecture_comp_dibs}

Figure \ref{fig:dibs_grid} presents the performance analysis of the DiBS method across various neural network configurations. As with the DCDI method, we evaluate models with different numbers of layers and hidden dimension sizes.
Consistent with DCDI, we find that the optimal performance for DiBS is achieved by a two-layer model with a hidden dimension of 4. However, the performance landscape for DiBS exhibits less variability across different model configurations. Single-layer models perform nearly as well as the optimal two-layer model.

Furthermore, we observe that more expressive models do not show a significant degradation in performance as was seen with DCDI. The overall differences in metric across all tested configurations are relatively small for DiBS, indicating a more consistent performance across varying levels of model complexity.

\begin{figure}[ht]
    \centering
    \includegraphics[width=1\linewidth]{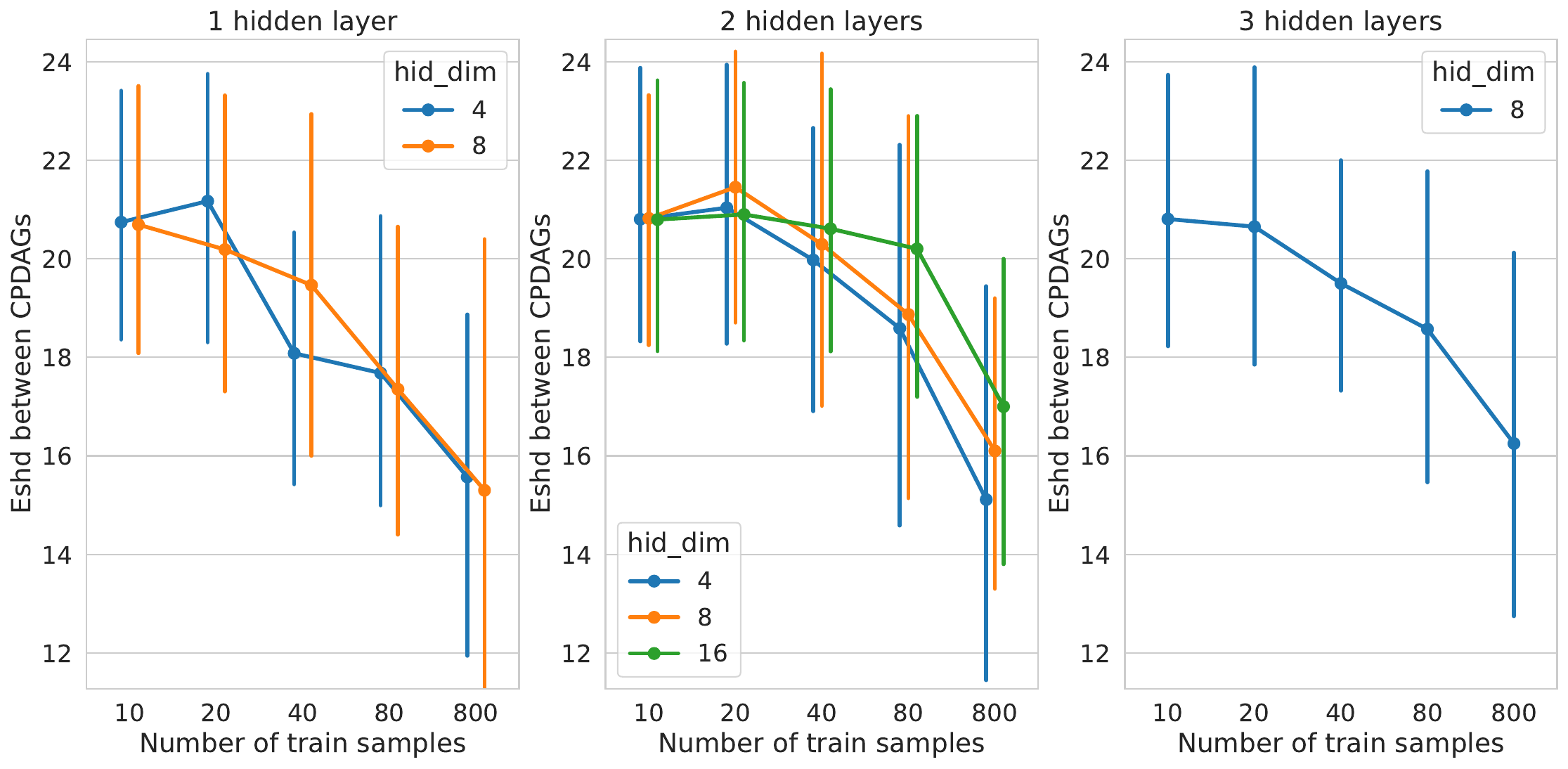}
    \caption{Comparison of the performance of DiBS depending on the model architecture and number of samples.}
    \label{fig:dibs_grid}
\end{figure}

\paragraph{BayesDAG}\label{model_architecture_comp_bayesdag}
Figure \ref{fig:bayes_grid} compares the performance of BayesDAG across different model architectures and sample sizes. For smaller sample sizes, BayesDAG's performance remains consistent, with noticeable differences emerging only at a sample size of 800. This suggests that BayesDAG requires more data to fully leverage its model capacity, unlike what we observed for DCDI and DiBS, where performance varied more significantly across sample sizes. Notably, the best-performing architecture for DiBS is a two-layer MLP with a hidden dimension of 4.

\begin{figure}[ht]
    \centering
    \includegraphics[width=1\linewidth]{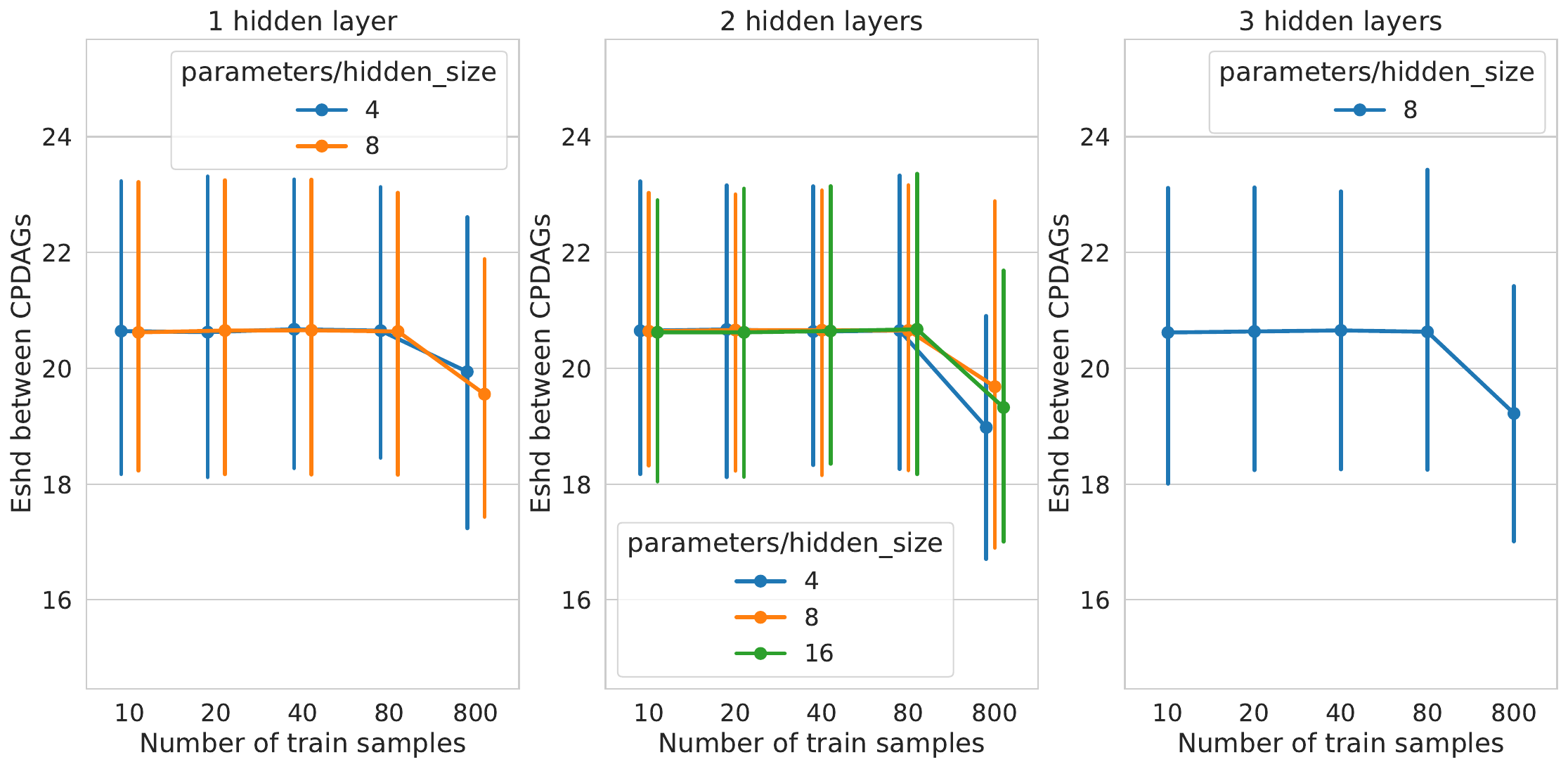}
    \caption{Comparison of the performance of DiBS depending on the model architecture and number of samples.}
    \label{fig:bayes_grid}
\end{figure}

\paragraph{SDCD}\label{model_architecture_comp_sdcd}

Figure \ref{fig:sdcd_grid} presents a similar comparison of SDCD performance across different MLP architectures and sample sizes. Interestingly, the three-layer architectures show stagnant performance regardless of sample size, while the one-layer models exhibit significant improvement as the sample size increases. Overall, the best performance is achieved with a one-layer MLP with 8 hidden units, although it remains comparable to the one-layer MLP with 4 hidden units and the two-layer MLP with 4 hidden units.

\begin{figure}[ht]
    \centering
    \includegraphics[width=1\linewidth]{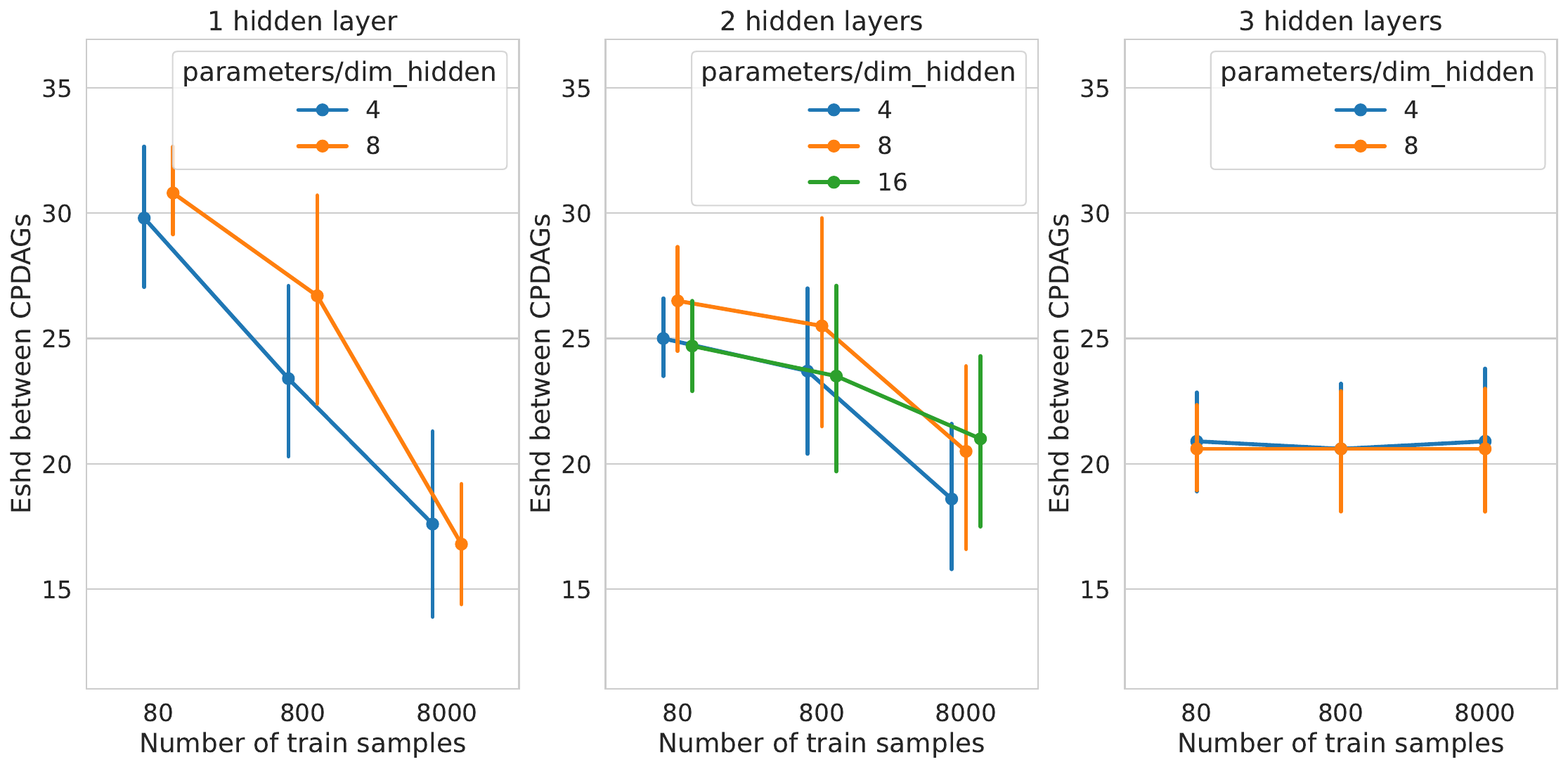}
    \caption{Comparison of the performance of SDCD depending on the model architecture and number of samples.}
    \label{fig:sdcd_grid}
\end{figure}

\paragraph{Model architecture}\label{model_options}
Inspired by BayesDAG, we also implemented layer normalization and residual connections to assess their impact. We conducted additional experiments on both the best-performing model ([4, 4]) and the largest model ([8, 8, 8]). The size of networks was similar to the one proposed in articles introducing tested methods: in DCDI it was [16, 16], for SDCD it was [10, 10], for DiBS [5, 5] and for BayesDAG it was a two layer network with a hidden size varying with dimensionality. The results of these tests are presented Figure~\ref{fig:model_options}. We show, there is no significant and consistent improvement across all networks, supporting our initial conclusion that variations in MLP architecture have minimal impact on performance.

\begin{figure}[ht]
    \centering
    \includegraphics[width=0.7\linewidth]{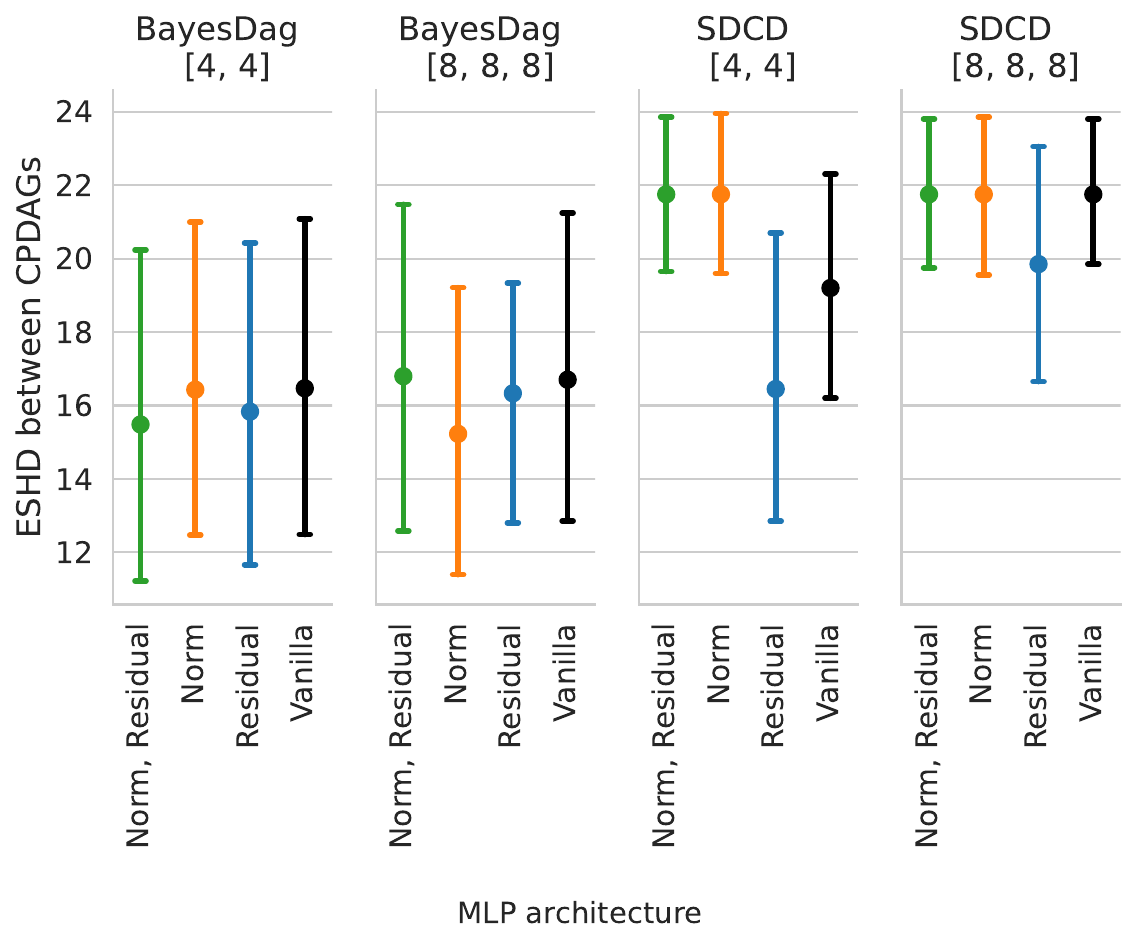}
    \caption{Comparison of the performance of SDCD depending on the model architecture and number of samples.}
    \label{fig:model_options}
\end{figure}

% \subsection{Benchmarking methods on graph with \ERfive} \label{appendix:p5e5}

% Table~\ref{tab:p5e5_main} shows the performance of the benchmarked methods on $\ERfive$ dataset, generated as described in Appendix~\ref{appendix:dataset_generation}.

% \begin{table}[ht]
%     \centering
%     \begin{tabular}{c|cc}
%          Method & $\eshdcpdag$ & $\fscorecpdag$  \\
%        \midrule
%         DCDI &  5.70 \tiny{(3.70, 8.10)} & 0.60 \tiny{(0.46, 0.74)} 	 \\
%        \midrule
%         BayesDAG & \textbf{3.90 \tiny{(3.63, 4.33)}} & \textbf{0.78 	\tiny{(0.77, 0.81)}} 	  \\
%        % \midrule
%         % DiBS &   \tiny{} & \\
%        \midrule
%         SDCD & 5.40 \tiny{(3.80, 6.70)} & 0.60 \tiny{(0.35, 0.69} \\
%     \end{tabular}
%     \caption{Results }
%     \label{tab:p5e5_main}
% \end{table}

% Trends in the presented results is slightly different, with BayesDAG performing the best in terms of $\eshdcpdag$ and $\fscorecpdag$, and having the smallest confidence interval. Nevertheless, all methods predict more than half edges wrong.

\section{Justification of evaluation metrics}
\label{app:metrics_justification}
We design metrics based on popular SHD, F1-score
% , and SID 
metrics, which we explain shortly below.

\paragraph{The Structural Hamming Distance.} SHD~\citep{shd} quantifies the difference between the predicted graph and the ground truth graph by counting the number of edge insertions, deletions, and reversals required to transform one into the other. SHD values indicate the degree of error in recovering the true causal structure: lower SHD values signify better predictions, while higher values indicate more significant discrepancies.

\paragraph{The F1-score.}
The F1-Score measures the harmonic mean of precision and recall for edge predictions, where precision reflects the fraction of correctly predicted edges among all predicted edges, and recall reflects the fraction of correctly predicted edges among the true edges.

% \paragraph{The Structural Intervention Distance.}  SID~\citep{Peters2015Structural} assesses the quality of a predicted causal graph by measuring the difference in predicted causal effects under interventions. The SID computes paths between all pairs of variables within the graphs and checks whether the causal relationships are preserved. Unlike SHD, which counts discrepancies based on edge presence or absence without regard for directionality or causal significance, SID provides a more nuanced evaluation that considers how interventions would affect the relationships depicted in the graphs.

We evaluate causal discovery methods based on observational data. In general, in this setup, it is only possible to recover true DAG up to a Markov Equivalence Class, a class of graphs with the same conditional independence relationships, due to identifiability issues TODO cite pearl?. If we were to compare the predicted and ground true graphs using standard metrics like SHD or F1-score we would obtain distorted results --- graphs from the MEC class do not generally receive these metrics' optimal values.

Therefore, we modify the formulation of the metrics to account for the limitations of causal discovery from observational data. We define $\eshdcpdag$ and $\fscorecpdag$. These metrics attain their optimal values, 0 and 1 correspondingly, for all DAG from ground truth MEC. Additionally, some of the benchmarked methods are Bayesian thus return the posterior over possible solutions. For those methods, we design metrics that compute the expected value over the posterior and approximate it with the Montecarlo estimator based on a sample of size 100.

We define \textbf{Expected SHD between CPDAGs} as:

\begin{equation} \eshdcpdag(\mathcal{G}, \sG) = \mathbb{E}_{\mathcal{G}^* \sim \sG} [\text{SHD}(\text{CPDAG}(\mathcal{G}), \text{CPDAG}(\mathcal{G}^*))], \end{equation}

where $\sG$ is the resulting distribution of graphs, $\mathcal{{G}^*}$ is a graph sampled from $\sG$ and $\mathcal{{G}}$ is the ground true graph.
% \paragraph{Expected F1-Score between CPDAGs}
% The F1-Score measures the harmonic mean of precision and recall for edge predictions, where precision reflects the fraction of correctly predicted edges among all predicted edges, and recall reflects the fraction of correctly predicted edges among the true edges.
Similarly, we compute the \textbf{Expected F1-Score between the CPDAGs}:

\begin{equation} \fscorecpdag(\mathcal{G}, \sG) = \mathbb{E}_{\mathcal{G}^* \sim \sG} [\text{F1-Score}(\text{CPDAG}(\mathcal{G}), \text{CPDAG}(\mathcal{G}^*))]. \end{equation}
subsection{Influence of sample samples on performance on the graph with \ERfive} \label{p5e5_samples}

Figure~\ref{fig:p5e5_samples} shows the $\eshdcpdag$ of benchmachmarked methods for different sample sizes. For all observational sample sizes, SDCD and DCDI have a large confidence interval. For datasets with 2,500 and 8,000 samples, BayesDAG performs better than other benchmarked methods, getting small confidence interval for 8,000 samples.

\begin{figure}
    \centering
    \includegraphics[width=0.8\linewidth]{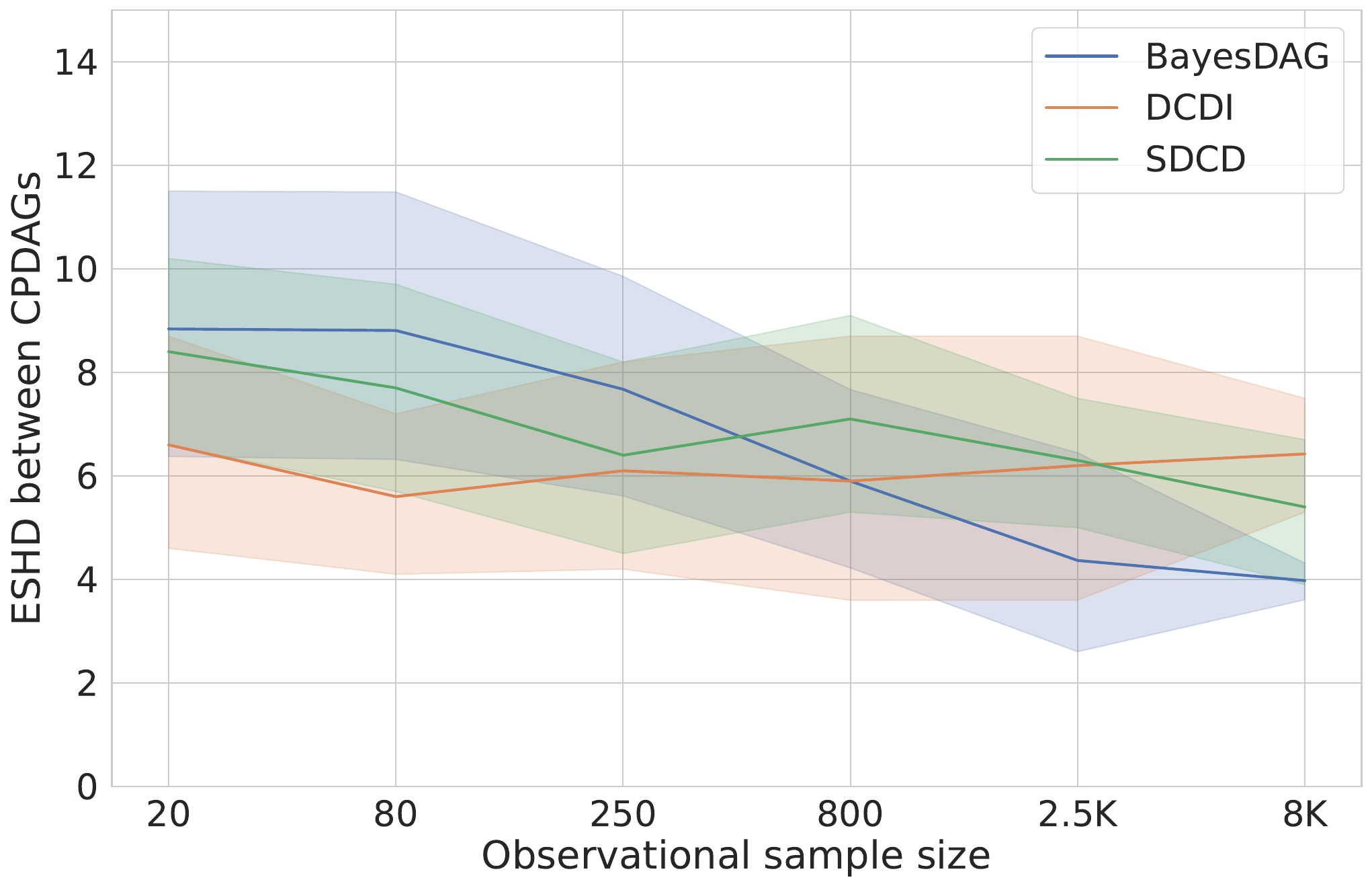}
    \caption{Comparision of $\eshdcpdag$ for benchmarked methods on $\ERfive$ dataset, averaged over 10 graphs.}
    \label{fig:p5e5_samples}
\end{figure}

\end{document}